\documentclass[letterpaper]{article} 
\usepackage{aaai2026}  
\usepackage{times}  
\usepackage{helvet}  
\usepackage{courier}  
\usepackage[hyphens]{url}  
\usepackage{graphicx} 
\usepackage{natbib}  
\usepackage{caption} 
\usepackage{algorithm}
\usepackage{algorithmic}

\usepackage{newfloat}
\usepackage{listings}
\DeclareCaptionStyle{ruled}{labelfont=normalfont,labelsep=colon,strut=off} 
\floatstyle{ruled}
\newfloat{listing}{tb}{lst}{}
\floatname{listing}{Listing}
\usepackage{graphics}
\usepackage{epsfig}
\usepackage{mathptmx}
\usepackage{times}
\usepackage{amsmath}
\usepackage{arydshln}
\usepackage{amssymb}
\usepackage{array}
\usepackage{graphicx}
\usepackage{subcaption}
\usepackage{adjustbox}
\usepackage{booktabs}
\usepackage[table]{xcolor}
\usepackage{multirow}

\newcommand{\eqnref}[1]{Eq.~(\ref{#1})}

\newcommand{\Figref}[1]{Figure~\ref{#1}}

\newcommand{\Tabref}[1]{Table~\ref{#1}}

\title{OceanSplat: Object-aware Gaussian Splatting with Trinocular View Consistency for Underwater Scene Reconstruction}
\author {
    Minseong Kweon\textsuperscript{\rm 1},
    Jinsun Park\textsuperscript{\rm 2}\footnote{Corresponding Author}
}
\affiliations {
    \textsuperscript{\rm 1}University of Minnesota, MN, USA\\
    \textsuperscript{\rm 2}Pusan National University, South Korea\\
    kweon021@umn.edu, jspark@pusan.ac.kr
}

\begin{document}

\maketitle

\begin{abstract}
We introduce OceanSplat, a novel 3D Gaussian Splatting-based approach for high-fidelity underwater scene reconstruction.
To overcome multi-view inconsistencies caused by scattering media, we design a trinocular setup for each camera pose by rendering from horizontally and vertically translated virtual viewpoints, enforcing view consistency to facilitate spatial optimization of 3D Gaussians.
Furthermore, we derive synthetic epipolar depth priors from the virtual viewpoints, which serve as self-supervised depth regularizers to compensate for the limited geometric cues in degraded underwater scenes.
We also propose a depth-aware alpha adjustment that modulates the opacity of 3D Gaussians during early training based on their depth along the viewing direction, deterring the formation of medium-induced primitives.
Our approach promotes the disentanglement of 3D Gaussians from the scattering medium through effective geometric constraints, enabling accurate representation of scene structure and significantly reducing floating artifacts.
Experiments on real-world underwater and simulated scenes demonstrate that OceanSplat substantially outperforms existing methods for both scene reconstruction and restoration in scattering media.
\end{abstract}

\begin{links}
    \link{Project Page}{https://oceansplat.github.io}
\end{links}

\section{Introduction}
\label{sec:intro}
Underwater scene reconstruction is essential for several marine robotics tasks, including seafloor mapping~\cite{kapoutsis2016real,liu2020underwater,joshi2022underwater}, ecological monitoring~\cite{marre2019monitoring,girdhar2023curee}, and subsea infrastructure inspection~\cite{fang2023integration,huang2025visual}. While visual data can generally support robotic operations effectively, the underwater optical properties, such as wavelength-dependent attenuation~\cite{akkaynak2017space}, scattering~\cite{mcglamery1980computer,jaffe2002computer,chen2021underwater}, and low illumination~\cite{marques2020l2uwe}, significantly degrade perceptual cues~\cite{yu2023task,zhang2024dcgf} and hinder the deployment of vision-based autonomous systems~\cite{de2021impact}. Consequently, achieving geometric understanding and spatial representation from this low-quality imagery is essential for practical operation of autonomous platforms. Although advances in remotely operated vehicles (ROVs) and autonomous underwater vehicles (AUVs) have greatly enabled the acquisition of large-scale underwater imagery~\cite{wynn2014autonomous}, leveraging this data for scene representation remains highly challenging due to adverse visual conditions~\cite{beall20103d,huang2025visual}.
\begin{figure}[t!]
    \centering
    \includegraphics[width=\linewidth]{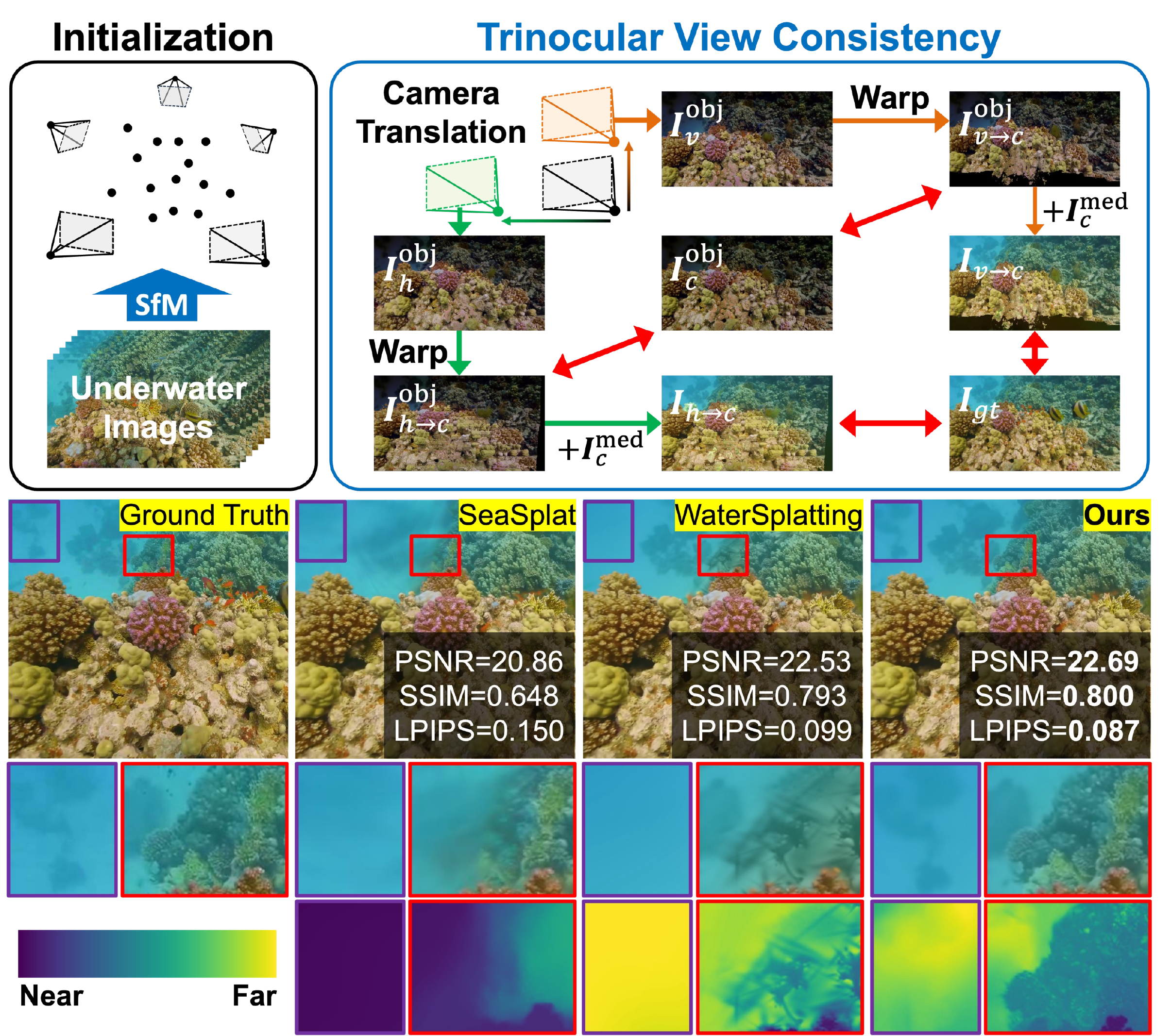} \\
    \caption{\footnotesize OceanSplat overcomes scattering and attenuation effects through trinocular view consistency, preserving object structure and enabling high-quality underwater 3D reconstruction.}
    \label{fig:teaser}
\end{figure}
To address these issues in underwater perception, the use of Neural Radiance Fields (NeRF)~\cite{mildenhall2020nerf} and 3D Gaussian Splatting (3DGS)~\cite{kerbl20233d}, which have shown strong capability in 3D reconstruction, has emerged as a promising direction. However, these methods rely on assumptions tailored to in-air image acquisition, motivating recent research to integrate underwater image formation model~\cite{akkaynak2019sea}.

NeRF-based methods~\cite{sethuraman2023waternerf, zhang2023beyond, levy2023seathru,ramazzina2023scatternerf,tang2024neural} extend their volumetric rendering framework by integrating light-transmitting physics and scattering phenomena.
However, these methods implicitly represent both object and backscatter components, which impedes accurate geometric representation and also results in slow rendering speed.
More recently, 3DGS-based methods~\cite{yang2024seasplat, li2024watersplatting, wang2024uw} have been adapted for underwater environments, benefiting from their superior rendering performance and fast training.
Despite notable progress in these studies, medium intensity is often absorbed into the 3D Gaussians, leading to numerous floating artifacts and degrading reconstruction quality.

Unlike previous approaches, our goal is to robustly recover scene geometry in underwater environments subject to scattering and attenuation, without depending on external geometric supervision.
To this end, we propose \textbf{OceanSplat}: \textbf{O}bje\textbf{c}t-awar\textbf{e} Gaussi\textbf{an} \textbf{Splat}ting for geometrically consistent underwater scene reconstruction.
Our method enforces trinocular view consistency across translated camera views along two orthogonal axes, resolving multi-view inconsistencies and effectively guiding 3D Gaussians to align with the scene structure.
Additionally, we derive a synthetic epipolar depth prior through triangulation between the translated viewpoints and leverage it for self-supervised depth regularization.
Furthermore, we apply a depth residual loss that constrains the $z$-component of each 3D Gaussian toward the alpha-blended depth to reduce floating artifacts.
We also propose a depth-aware alpha adjustment that regulates the transparency of 3D Gaussians during the early stages of training based on their $z$-component and viewing direction, thereby discouraging the formation of medium-induced primitives.
As shown in \Figref{fig:teaser}, our method improves geometric accuracy for object representation under challenging scattering conditions by preventing the entanglement of 3D Gaussians with the scattering medium. Our key contributions are outlined in the following points:
\begin{itemize}
\item We propose OceanSplat, a novel underwater scene reconstruction method leveraging 3D Gaussian Splatting, which employs trinocular view consistency to align 3D Gaussians with scene geometry in scattering media.
\item We introduce synthetic epipolar depth priors derived from translated virtual viewpoints via triangulation, enabling self-supervised depth regularization.
\item We propose a depth-aware alpha adjustment that regulates the opacity of 3D Gaussians during the early training stage, suppressing medium-induced primitives.
\item Our approach surpasses existing state-of-the-art methods in both qualitative and quantitative evaluations across real-world and simulated datasets.

\end{itemize}

\begin{figure*}[t]
    \centering
    \includegraphics[width=0.96\linewidth]{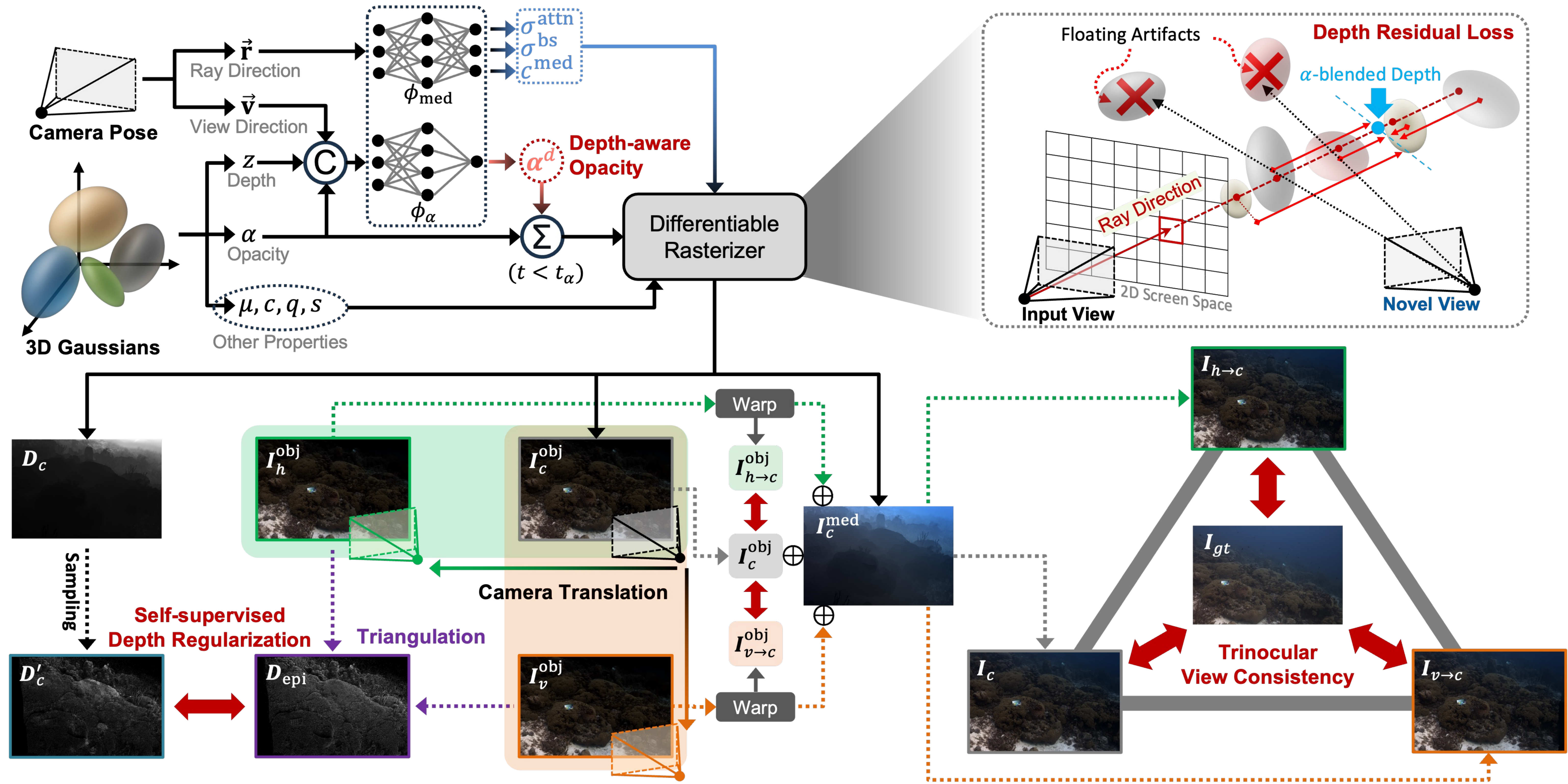} \\
    \caption{Overview of OceanSplat. We enforce trinocular view consistency by inverse warping rendered images from two translated camera poses to guide the spatial optimization of 3D Gaussians. From these poses, we derive a synthetic epipolar depth prior via triangulation, which provides self-supervised geometric constraints. Additionally, depth-aware alpha adjustment suppresses erroneous 3D Gaussians early and aligns rendered depth with the Gaussian $z$-component to prevent floating artifacts.}
    \label{fig:main_architecture}
\end{figure*}

\section{Related Work}
\label{sec:related_work}

\subsubsection{Underwater Scene Reconstruction}
Underwater scene reconstruction aims to represent precise scene structure and restore true color despite severe light attenuation and scattering. Initial approaches jointly estimated medium-dependent attenuation during reconstruction~\cite{skinner2017automatic} and exploited global context with active labeling for sonar-based modeling~\cite{debortoli2018real}. Subsequent works~\cite{sethuraman2023waternerf,zhang2023beyond,tang2024neural} embed physics-based underwater light transport and scattering into NeRF~\cite{mildenhall2020nerf} to jointly recover geometry and true color appearance. Seathru-NeRF~\cite{levy2023seathru} leverages the \textit{SeaThru} image formation model~\cite{akkaynak2019sea} within NeRF to successfully achieve novel view synthesis in real-world underwater scenes. ScatterNeRF~\cite{ramazzina2023scatternerf} uses inverse rendering, while DecoNeRF~\cite{zhang2025decoupling} employs pseudo-labeling to disentangle scattering media from scene content and restore clear views in foggy scenes. Although these methods significantly improve reconstruction quality, they are often hindered by slow rendering speed and high memory consumption. Recently, 3DGS~\cite{kerbl20233d} has gained increasing attention for underwater scene reconstruction. SeaSplat~\cite{yang2024seasplat} incorporates underwater physics into 3D Gaussians to improve color restoration and structural fidelity, while WaterSplatting~\cite{li2024watersplatting} combines implicit medium modeling with explicit object representation for efficient, high-quality reconstruction. UW-GS~\cite{wang2024uw} further enhances geometry using pseudo-depth priors from foundation models~\cite{yang2024depth}. 
Despite these advances, existing 3DGS-based methods have difficulty recovering accurate scene structure in underwater environments.

In contrast, our method enforces a trinocular view consistency constraint that enhances the spatial coherence of 3D Gaussians, and leverages self-supervised depth regularization based on epipolar geometry to achieve geometrically accurate underwater scene reconstruction.

\subsubsection{Multi-View Stereo} Multi-view stereo (MVS) algorithms reconstruct detailed 3D geometry by leveraging images acquired from multiple calibrated viewpoints. Early approaches employed voxelization~\cite{kutulakos2000theory}, dense point clouds~\cite{furukawa2009accurate}, and depth maps~\cite{galliani2015massively,schonberger2016structure}. In the deep learning era, MVSNet~\cite{yao2018mvsnet} introduced cost volumes to form geometry-aware 3D representations, greatly enhancing accuracy. Building upon these advancements, MVS has been integrated into neural rendering~\cite{mildenhall2020nerf,kerbl20233d}, where NeRF-based methods~\cite{chen2021mvsnerf,wang2021neus,fu2022geo,wu2025sparis} utilize these concepts for surface reconstruction, while 3DGS-based approaches~\cite{chen2023neusg,guedon2024sugar,huang20242d,chung2024depth,chen2024pgsr,chen2024mvsplat,liu2024mvsgaussian,li2024dngaussian,huang2025fatesgs} harness multi-view relationships to enhance geometric fidelity. Among these, recent efforts~\cite{safadoust2024self,han2024binocular} have demonstrated promising results by constructing virtual stereo pairs to enforce binocular stereo consistency, effectively regularizing 3D Gaussians.

To better address geometric ambiguities inherent to scattering media, we extend binocular stereo vision to a trinocular configuration by introducing a vertically translated camera, facilitating spatial optimization of 3D Gaussians.

\section{Preliminaries}
\label{sec:pre}
\subsection{3D Gaussian Splatting}
3D Gaussian Splatting~\cite{kerbl20233d} models a scene with anisotropic 3D Gaussian primitives, each parameterized by a center position \( \mu \in \mathbb{R}^3 \), covariance matrix \( \Sigma \in \mathbb{R}^{3 \times 3} \), color \( c \), opacity \( \alpha \), and view-dependent appearance modeled by spherical harmonics. Each 3D Gaussian defines its spatial influence at a point \( \mathbf{X} \in \mathbb{R}^3 \) centered at \( \mu \) as:
\begin{equation}
G(\mathbf{X}) = e^{-\tfrac{1}{2}(\mathbf{X} - \mu)^\top \Sigma^{-1}(\mathbf{X} - \mu)}.
\label{eq:3dgs_}
\end{equation}
For rendering, 3D Gaussians are projected onto 2D screen-space and composited via alpha-blending of \( N \) primitives that are sorted by depth and overlap at each pixel:
\begin{equation}
C = \sum_{i=1}^{N} T_i \cdot \alpha_i \cdot c_i, \quad T_i = \prod_{j=1}^{i-1}(1 - \alpha_j),
\label{eq:3dgs}
\end{equation}
where \( C \), \( N \), and \( T_i \) denote the rendered color, the number of 3D Gaussians intersected by the ray, and the transmittance, respectively. Then, 3D Gaussians are optimized via an \text{L1} loss combined with \text{D-SSIM} on the rendered images.

\subsection{Underwater Image Formation Model}
To render a 3D underwater scene, the image formation is guided by the revised underwater image formation model~\cite{akkaynak2018revised}, which decomposes the observed image into a direct component attenuated by the medium with respect to depth $z$, and a backscatter component accumulated along the viewing direction, as follows:
\begin{equation}
C = C^{\text{obj}} \cdot e^{-\sigma^{\text{attn}} \cdot z} + C^{\infty} \cdot \left(1 - e^{-\sigma^{\text{bs}} \cdot z}\right),
\label{eq:uw_formation}
\end{equation}
where $C^{\text{obj}}$, $C^{\infty}$, $\sigma^{\text{attn}}$, and $\sigma^{\text{bs}}$ denote the unattenuated object color, the asymptotic backscatter color, the attenuation coefficient, and the backscattering coefficient, respectively.

\section{Methodology}
\label{sec:method}
\Figref{fig:main_architecture} shows the overall architecture of our method. We begin by initializing 3D Gaussians from a collection of RGB images \( I \in \mathbb{R}^{H \times W \times 3} \), along with their corresponding camera intrinsics \( K \in \mathbb{R}^{3 \times 3} \) and extrinsic parameters \( P \in \mathbb{R}^{4 \times 4} \), which represent the homogeneous transformation matrix, recovered using Structure-from-Motion (SfM)~\cite{schonberger2016structure}. We model medium properties along the ray $\vec{\mathbf{r}}$ using an MLP \( \phi_{\text{med}} \), following~\citep{li2024watersplatting}, which directly predicts attenuation \( \sigma^{\text{attn}} \), backscattering \( \sigma^{\text{bs}} \), and medium color \( c^{\text{med}} \) as follows:
\begin{equation}
\sigma^{\text{attn}},\ \sigma^{\text{bs}},\ c^{\text{med}} = \phi_{\text{med}}(\vec{\mathbf{r}}).
\end{equation}

Subsequently, by integrating \eqnref{eq:3dgs} with \eqnref{eq:uw_formation}, the contributed colors of the objects and medium along the ray in underwater scenes are formulated as follows:
\begin{equation}
C^{\text{obj}} = \sum_{i=1}^{N} T^{\text{obj}}_i \cdot \alpha_i \, \cdot c_i \, \cdot e^{-\sigma^{\text{attn}} z_i}, T^{\text{obj}}_i = \prod_{j=1}^{i-1}(1 - \alpha_j),
\end{equation}
\begin{equation}
C^{\infty} = T^{\text{obj}}_N \cdot c_{\text{med}} \cdot e^{-\sigma^{\text{bs}} z_N},
\end{equation}
\begin{equation}
C^{\text{med}} = \sum_{i=1}^{N} \left[T^{\text{obj}}_i \cdot c_{\text{med}} \cdot (e^{-\sigma^{\text{bs}} z_{i-1}} - e^{-\sigma^{\text{bs}} z_i})\right] + C^{\infty},
\end{equation}
where \( C^{\text{obj}} \), \( C^{\text{med}} \), \( C^{\infty} \), and \( T^{\text{obj}} \) denote the pixel-wise colors of the object, medium, background medium, and represent the transmittance of 3D Gaussians, respectively. Here, the object refers to physical structures submerged in water, while the medium represents the water itself. Based on this rendering framework, we describe our proposed methods for robust 3DGS-based underwater scene reconstruction.

\subsection{Trinocular View Consistency}
\label{subsec:trinocular}

\begin{figure}[t!]
    \centering
    \includegraphics[width=0.85\linewidth]{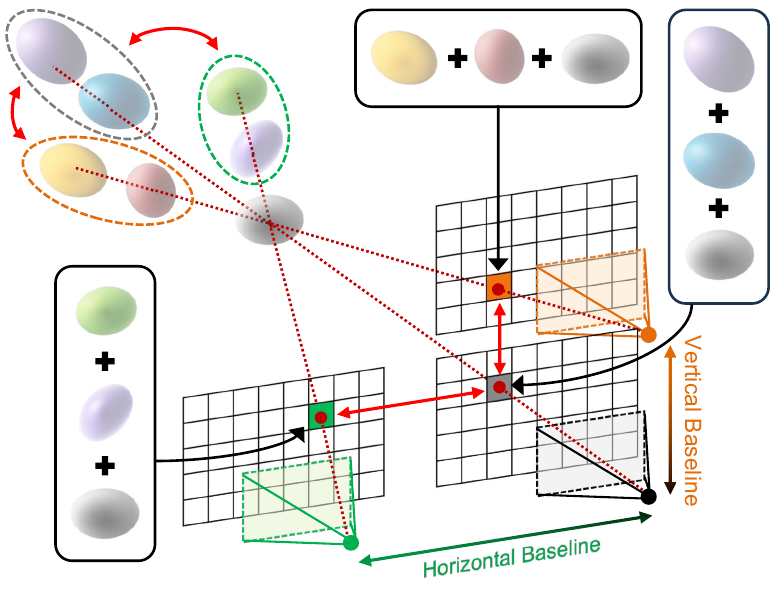} \\
    \caption{Pixel-wise color contributions across translated camera views in 3D Gaussian Splatting.}
    \label{fig:trinocular_1}
\end{figure}

We propose a trinocular stereo framework for 3DGS to improve object representation in scattering media. While prior works~\cite{safadoust2024self, han2024binocular} use only binocular stereo consistency, we show that a vertically translated viewpoint provides orthogonal constraints that complement horizontal stereo for 3D Gaussians. As shown in \Figref{fig:trinocular_1}, view-dependent sampling in alpha-blending often causes inconsistencies under minor viewpoint changes. Furthermore, previous stereo matching studies~\cite{okutomi1993multiple, imran2020unsupervised, shamsafar2021tristereonet} have shown that multi-baseline stereo surpasses single-baseline methods in accuracy and robustness. Inspired by these observations, we enforce trinocular view consistency with orthogonal baselines of varying lengths to better regularize the spatial position of 3D Gaussians.

From a given camera pose \( P_c \), we obtain the rasterized object and medium components, denoted by \( I^{\text{obj}}_c \) and \( I^{\text{med}}_c \), respectively. We define two virtual viewpoints by translating \( P_c \) along the horizontal $h$ and vertical $v$ axes as:
\begin{equation}
P_h = 
\begin{bmatrix}
\mathbb{I} &  \textbf{t}_h \\
\textbf{0}^{\top} & 1
\end{bmatrix}
P_c, \quad
P_v = 
\begin{bmatrix}
\mathbb{I} & \textbf{t}_v \\
\textbf{0}^{\top} & 1
\end{bmatrix}
P_c,
\end{equation}
where $\mathbb{I}$ is the $3\times3$ identity matrix, $\mathbf{t}_h = (b_h, 0, 0)^\top$ and $\mathbf{t}_v = (0, b_v, 0)^\top$ are translation vectors defining the horizontally and vertically translated camera poses $P_h$ and $P_v$, respectively, with $b_h, b_v \in \mathbb{R}$ denoting the baseline distances. Using these poses, we render two images \( I^{\text{obj}}_h \) and \( I^{\text{obj}}_v \) from viewpoints \( P_h \) and \( P_v \). We then apply disparity-based inverse warping to the rendered images. Let \( D_c \in \mathbb{R}^{H \times W \times 1} \) be the rendered depth map from \( P_c \). Based on stereo geometry, the horizontal and vertical disparity maps are computed as:
\begin{equation}
d_h(x,y) = \frac{f_h \cdot b_h}{D_c(x,y)}, \quad d_v(x,y) = \frac{f_v \cdot b_v}{D_c(x,y)},
\end{equation}
where \( f_h, f_v \in \mathbb{R} \) denote the focal lengths in the horizontal and vertical directions. The disparity maps \( d_h, d_v \in \mathbb{R}^{H \times W \times 1} \) are used to align the images rendered from the translated virtual viewpoints to the original view, and \( (x,y) \) denotes the spatial pixel coordinates. Then, the rendered images from the virtual viewpoints, $I^{\text{obj}}_h$ and $I^{\text{obj}}_v$, are inverse warped using disparity maps \( d_h \) and \( d_v \), respectively, in the direction opposite to the virtual camera shift, aligning them with the central image rendered from \( P_c \). The resulting inverse warped images \( I^{\text{obj}}_{h \rightarrow c}, I^{\text{obj}}_{v \rightarrow c} \in \mathbb{R}^{H \times W \times 3} \) are defined as:
\begin{equation}
\left\{
\begin{array}{l}
I^{\text{obj}}_{h \rightarrow c}(x, y) = I^{\text{obj}}_h(x - d_{h}(x,y),\ y), \\
I^{\text{obj}}_{v \rightarrow c}(x, y) = I^{\text{obj}}_v(x,\ y - d_{v}(x,y)),
\end{array}
\right.
\end{equation}
where \( d_{h}, d_{v}\in \mathbb{R}^{H \times W} \) denote the horizontal and vertical disparities.
To reconstruct the complete scene, the shifted object images are composited with the medium component \( I^{\text{med}}_c \) as:
\begin{equation}
\left\{
\begin{array}{l}
I_c = I^{\text{obj}}_c + I^{\text{med}}_c, \\
I_{h \rightarrow c} = I^{\text{obj}}_{h \rightarrow c} + I^{\text{med}}_c, \\
I_{v \rightarrow c} = I^{\text{obj}}_{v \rightarrow c} + I^{\text{med}}_c.
\end{array}
\right.
\end{equation}

Afterward, to emphasize dark regions as perceived in human vision, we follow prior studies~\cite{mildenhall2022nerf,levy2023seathru,li2024watersplatting}, and employ a regularized L1 loss $L_{\text{R-L1}}$ defined as:
\begin{equation}
L_{\text{R-L1}}(I_{1}, I_{2}) = \frac{1}{HW} \sum_{x,y} \left| \frac{I_{1}(x,y) - I_{2}(x,y) }{\mathsf{sg}(I_c(x, y)) + \epsilon}\right|,
\end{equation}
where $\mathsf{sg}(\cdot)$ and $\epsilon$ denote the stop-gradient operator and a small constant for numerical stability, respectively. This loss enforces object consistency across rendered images and photometric alignment with the ground truth $I_{gt}$ as follows:
\begin{equation}
L_{\text{obj-stereo}} = L_{\text{R-L1}}(I^{\text{obj}}_{h \rightarrow c}, I^{\text{obj}}_c) + L_{\text{R-L1}}(I^{\text{obj}}_{v \rightarrow c}, I^{\text{obj}}_c), \end{equation}
\begin{equation}
L_{\text{full-stereo}} = L_{\text{R-L1}}(I_{h \rightarrow c}, I_{gt}) + L_{\text{R-L1}}(I_{v \rightarrow c}, I_{gt}).
\end{equation}
In addition, we define the smoothness loss for horizontal and vertical disparities as follows:
\begin{equation}
L_{\text{smooth}}
=
\frac{1}{HW}
\sum_{d \in \{d_h, d_v\}}
\sum_{x,y}
\sum_{k \in \{h, v\}}
\left|\nabla_k d\right| \cdot e^{-\gamma \left|\nabla_k I_{gt}\right|},
\end{equation}
where $\nabla_{k}$ denotes the directional gradient along $k$-axis and $\gamma$ is a weighting factor, respectively. Finally, the overall trinocular view consistency loss is defined as:
\begin{equation}
L_{\text{tri}} = L_{\text{obj-stereo}} + L_{\text{full-stereo}} + L_{\text{smooth}}.
\end{equation}

\begin{figure*}[t!]
\centering
\renewcommand{\arraystretch}{0.5}
\setlength{\tabcolsep}{1pt}
\begin{tabular}{*{6}{>{\centering\arraybackslash}m{0.15\textwidth}}}
\text{\small Ground Truth} & \text{\small SeaThru-NeRF} & \text{\small 3DGS} & \text{\small SeaSplat} & \text{\small WaterSplatting} & \textbf{\small Ours} \\
\end{tabular}
\begin{tabular}{*{6}{>{\centering\arraybackslash}m{0.15\textwidth}}}
\includegraphics[width=0.15\textwidth]{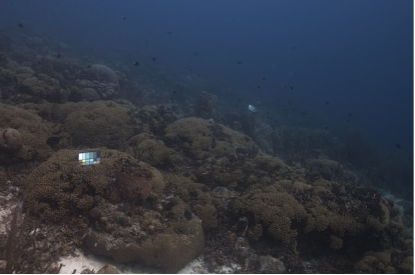} &
\includegraphics[width=0.15\textwidth]{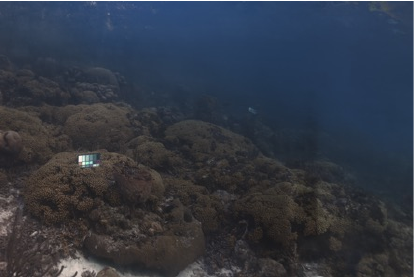} &
\includegraphics[width=0.15\textwidth]{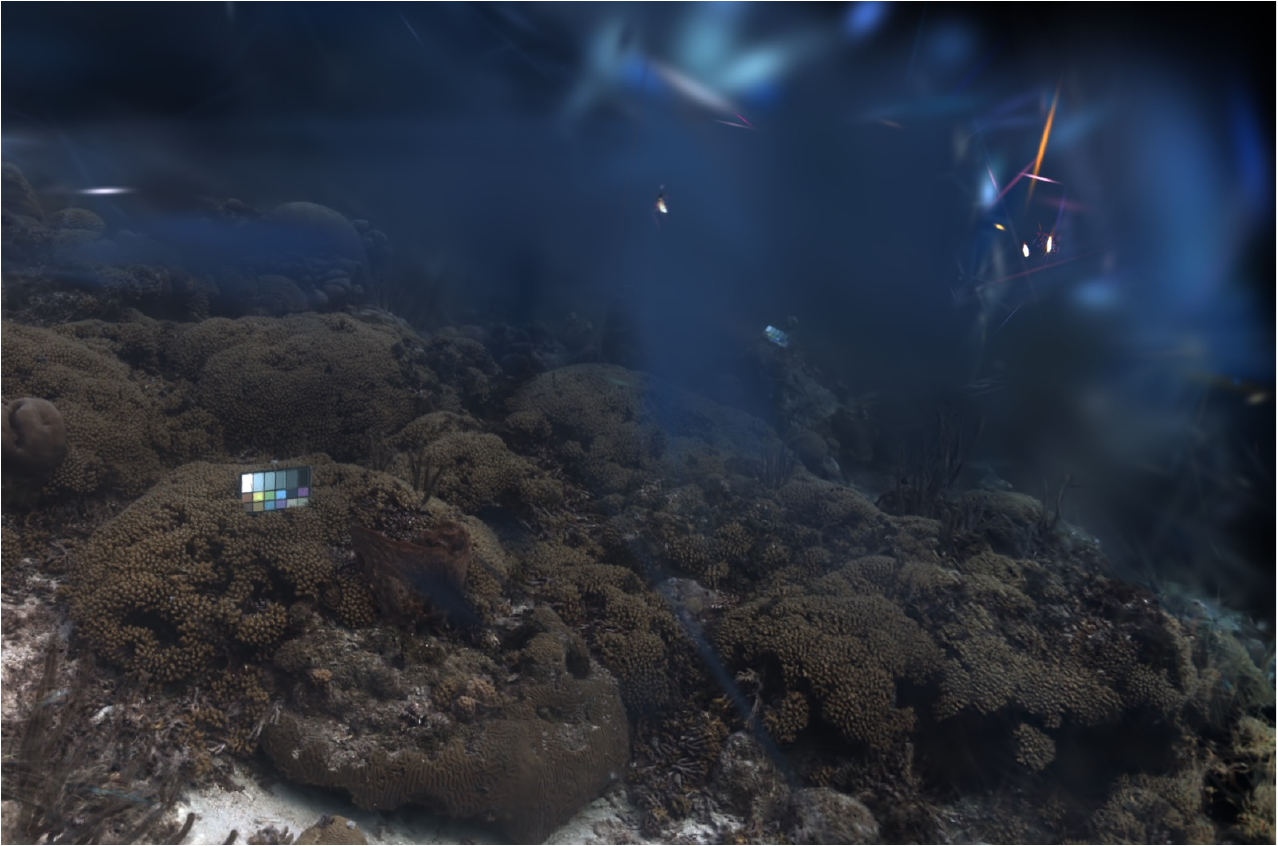} &
\includegraphics[width=0.15\textwidth]{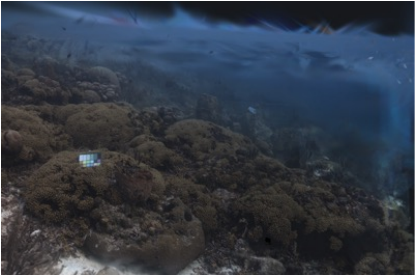} &
\includegraphics[width=0.15\textwidth]{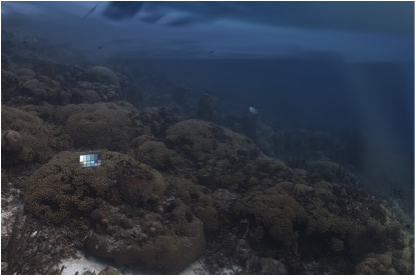} &
\includegraphics[width=0.15\textwidth]{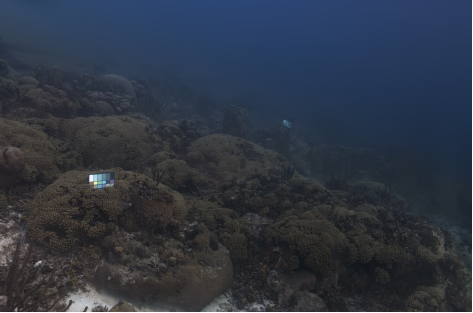} \\
\end{tabular}

\begin{tabular}{*{6}{>{\centering\arraybackslash}m{0.15\textwidth}}}
\includegraphics[width=0.15\textwidth]{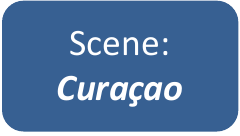} &
\includegraphics[width=0.15\textwidth]{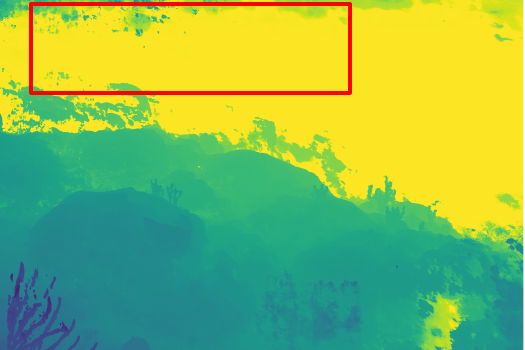} &
\includegraphics[width=0.15\textwidth]{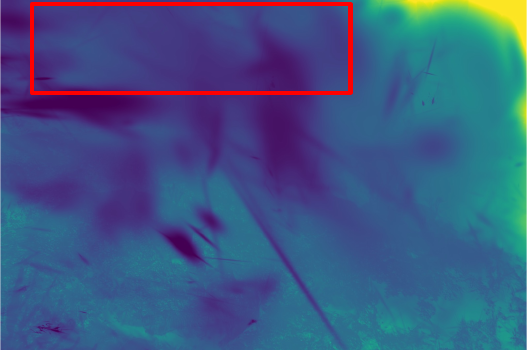} &
\includegraphics[width=0.15\textwidth]{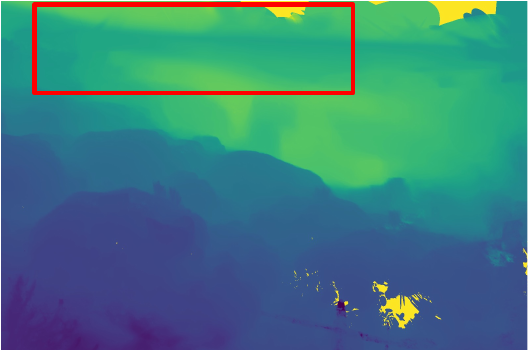} &
\includegraphics[width=0.15\textwidth]{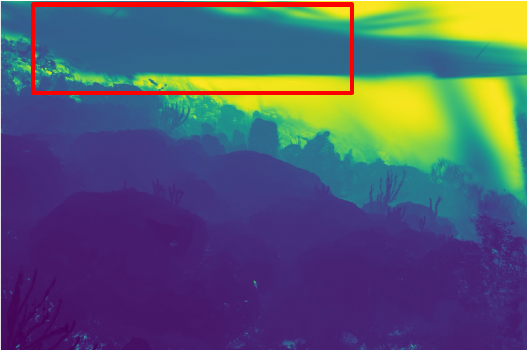} &
\includegraphics[width=0.15\textwidth]{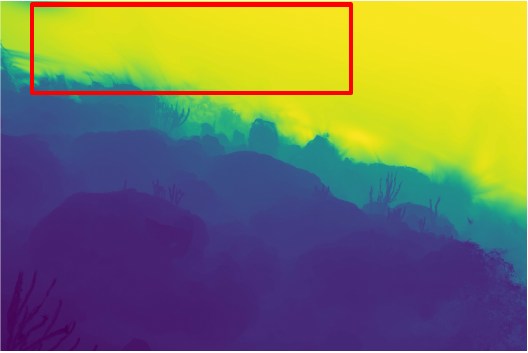} \\
\end{tabular}

\begin{tabular}{*{6}{>{\centering\arraybackslash}m{0.15\textwidth}}}
\includegraphics[width=0.15\textwidth]{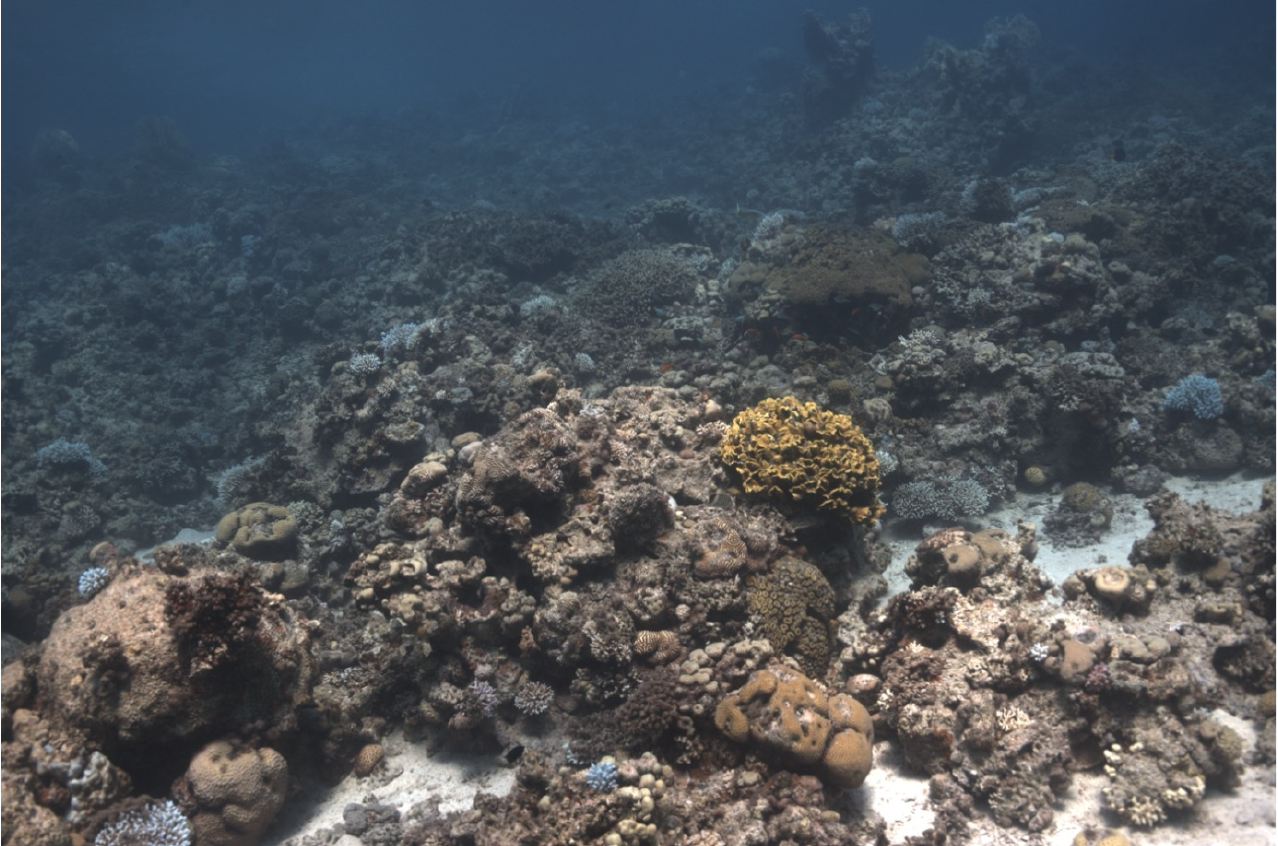} &
\includegraphics[width=0.15\textwidth]{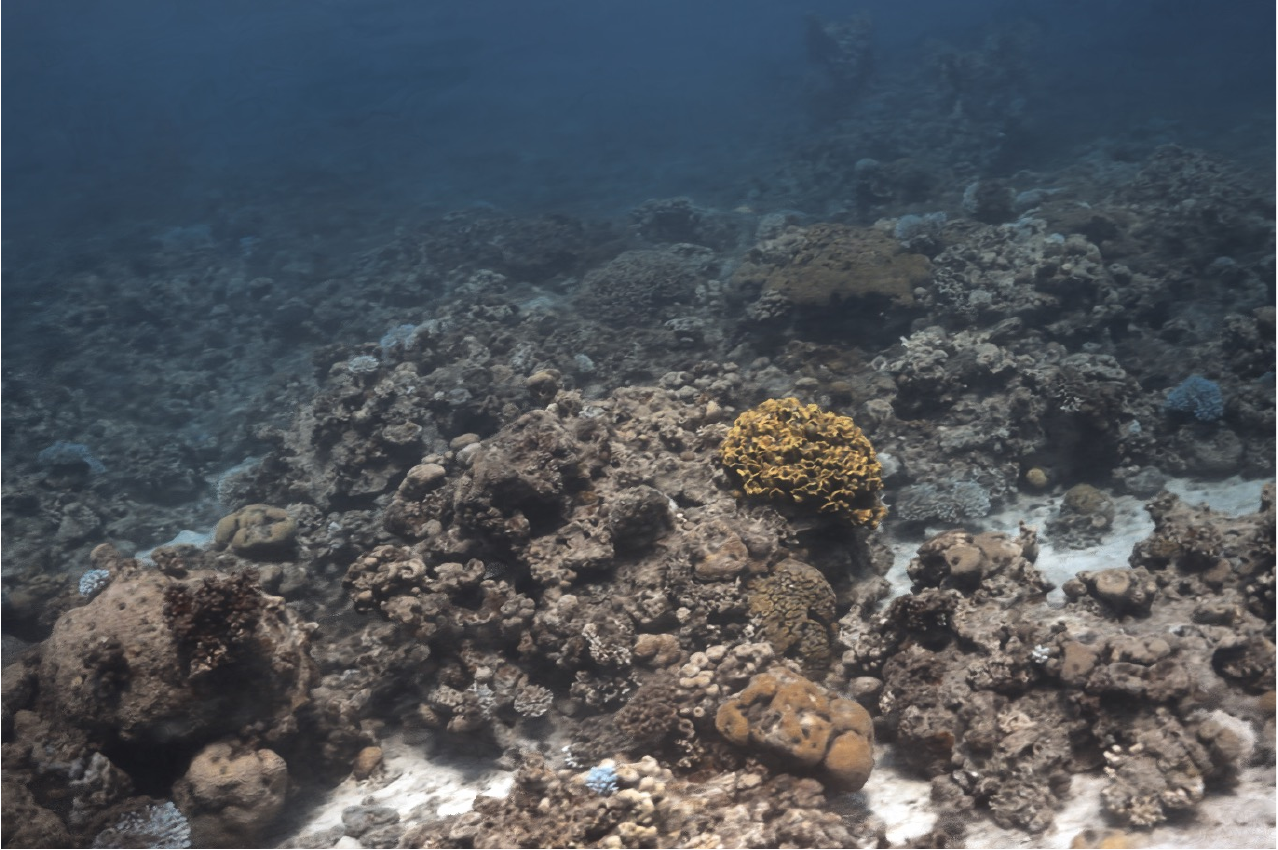} &
\includegraphics[width=0.15\textwidth]{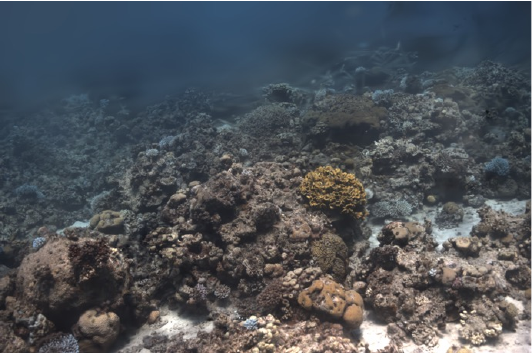} &
\includegraphics[width=0.15\textwidth]{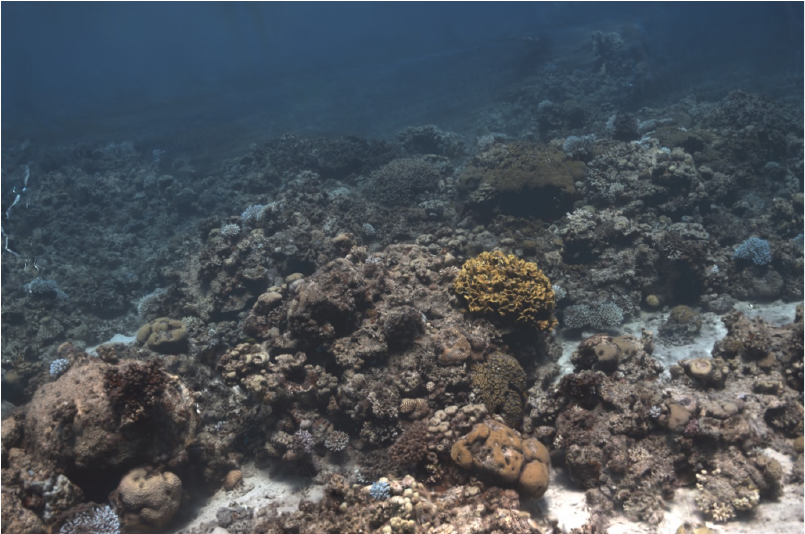} &
\includegraphics[width=0.15\textwidth]{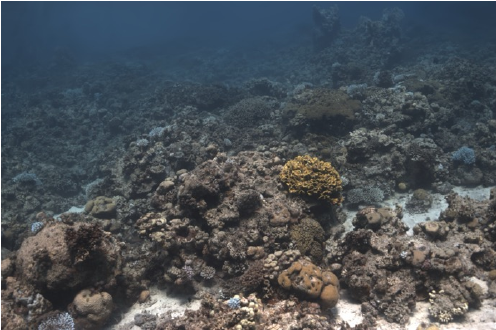} &
\includegraphics[width=0.15\textwidth]{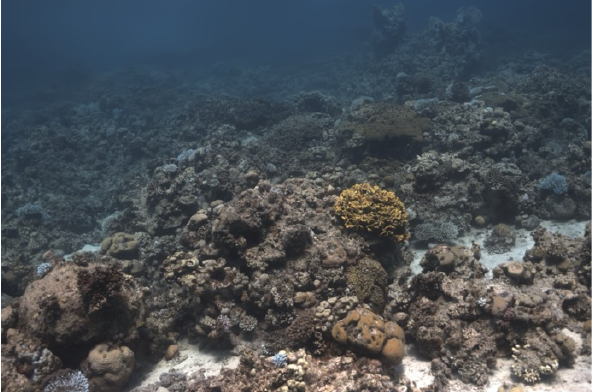} \\
\end{tabular}

\begin{tabular}{*{6}{>{\centering\arraybackslash}m{0.15\textwidth}}}
\includegraphics[width=0.15\textwidth]{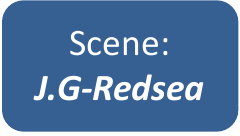} &
\includegraphics[width=0.15\textwidth]{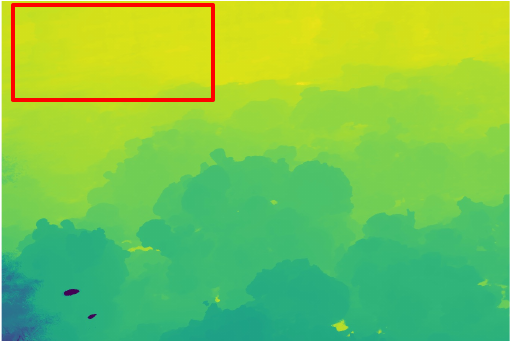} &
\includegraphics[width=0.15\textwidth]{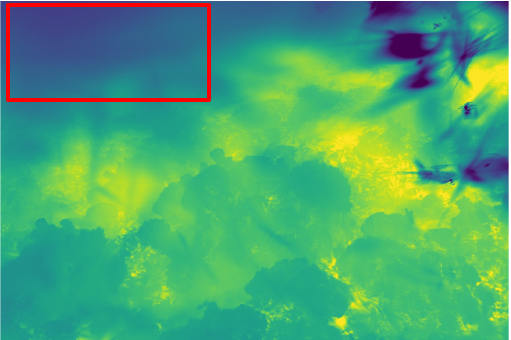} &
\includegraphics[width=0.15\textwidth]{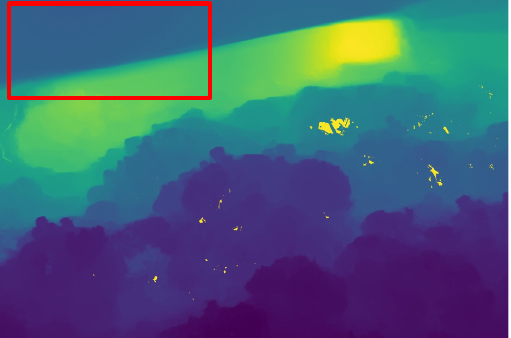} &
\includegraphics[width=0.15\textwidth]{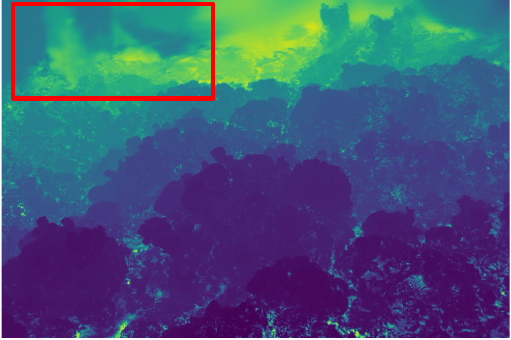} &
\includegraphics[width=0.15\textwidth]{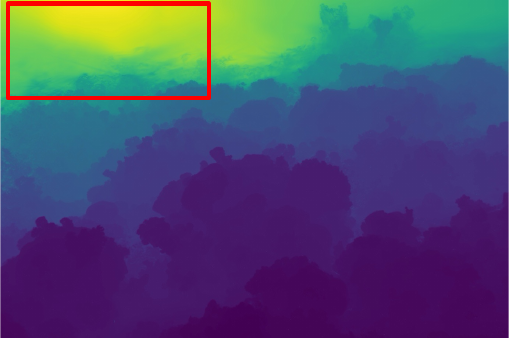} \\
\end{tabular}

\begin{tabular}{*{6}{>{\centering\arraybackslash}m{0.15\textwidth}}}
\includegraphics[width=0.15\textwidth]{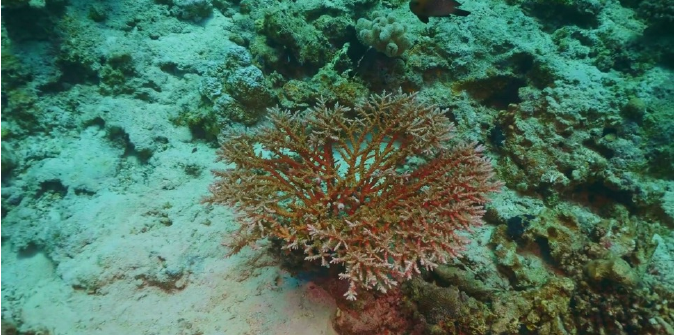} &
\includegraphics[width=0.15\textwidth]{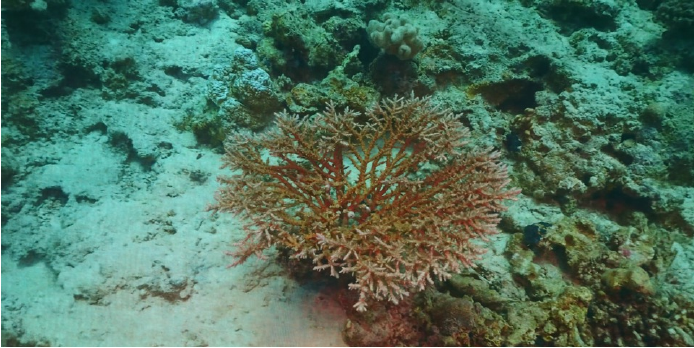} &
\includegraphics[width=0.15\textwidth]{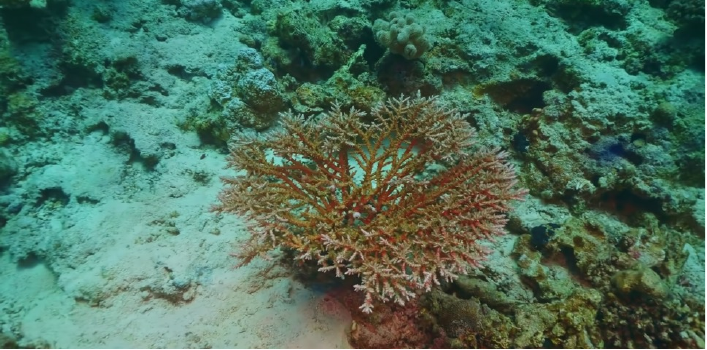} &
\includegraphics[width=0.15\textwidth]{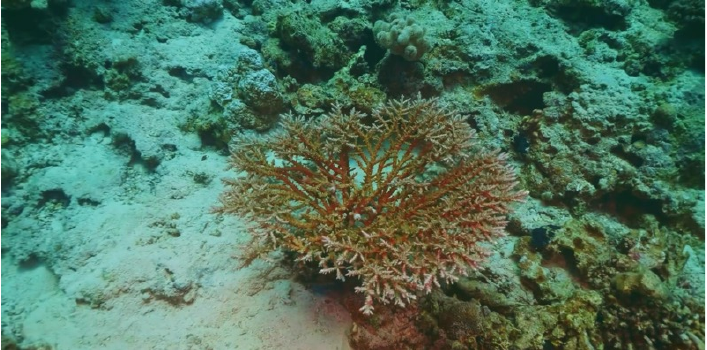} &
\includegraphics[width=0.15\textwidth]{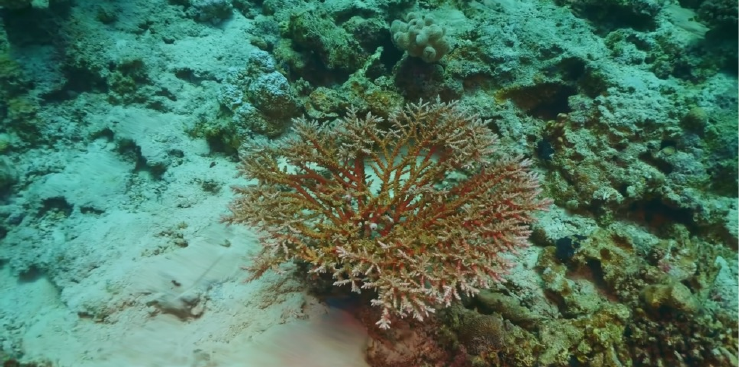} &
\includegraphics[width=0.15\textwidth]{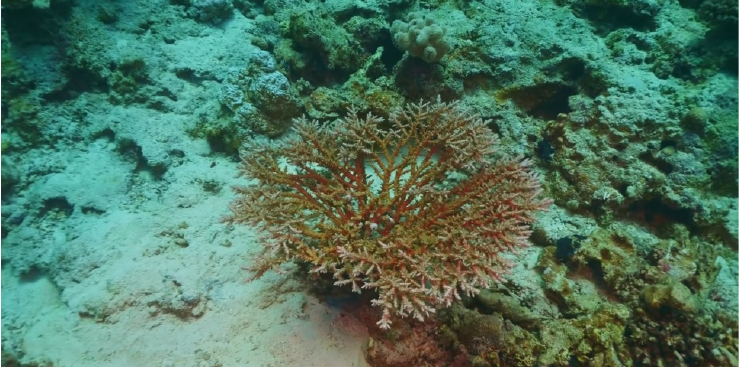} \\
\end{tabular}

\begin{tabular}{*{6}{>{\centering\arraybackslash}m{0.15\textwidth}}}
\includegraphics[width=0.15\textwidth]{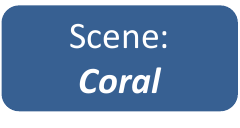} &
\includegraphics[width=0.15\textwidth]{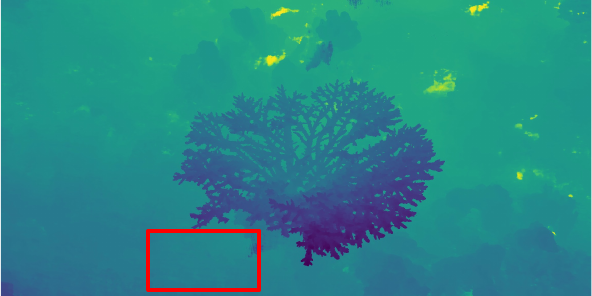} &
\includegraphics[width=0.15\textwidth]{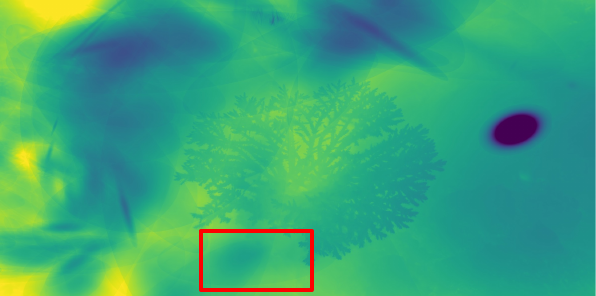} &
\includegraphics[width=0.15\textwidth]{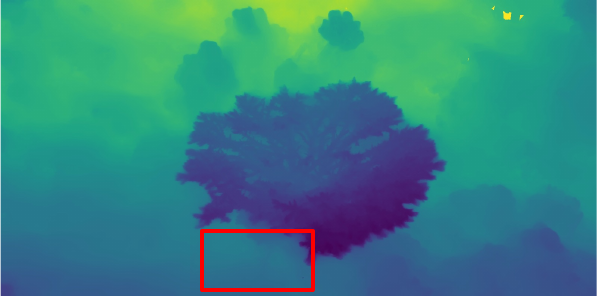} &
\includegraphics[width=0.15\textwidth]{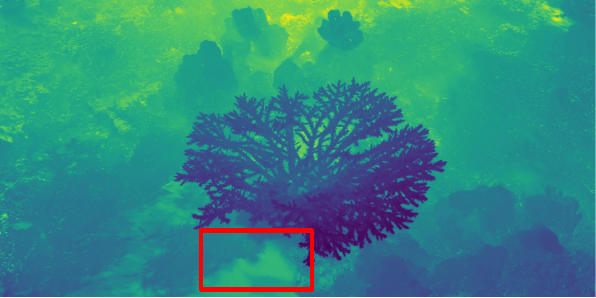} &
\includegraphics[width=0.15\textwidth]{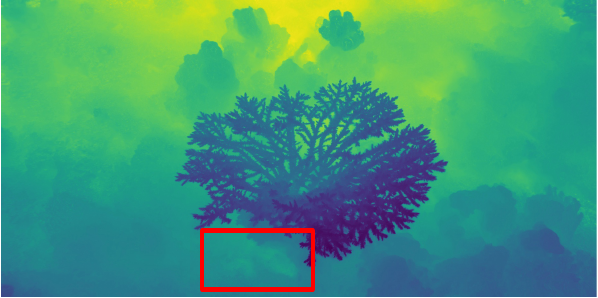} \\
\end{tabular}

\caption{Qualitative evaluation of novel view synthesis on diverse real-world underwater 3D scenes.}
\label{fig:qualitative_comparison}
\end{figure*}

\subsection{Synthetic Epipolar Depth Prior}
To enhance geometric alignment, we introduce a synthetic epipolar depth prior \( D_{\text{epi}} \), derived via triangulation between the translated viewpoints. For reliable correspondences in triangulation, we use 3D Gaussians sampled from the intersection of the view frusta of all trinocular camera views, with opacity above \( \tau_\alpha \). Each selected Gaussian is projected onto the image planes of \( P_h \) and \( P_v \) via projection matrices \( \mathbf{M}_h, \mathbf{M}_v \in \mathbb{R}^{3 \times 4} \), producing screen-space points \( \mathbf{x}_h^i, \mathbf{x}_v^i \in \mathbb{R}^2 \). Let the homogeneous coordinates be \( \tilde{\mathbf{x}}_k^i = [\mathbf{x}_k^{i\top}, 1]^\top \in \mathbb{R}^3 \) and \( \tilde{\mathbf{X}}_i = [\mathbf{X}_i^{\top}, 1]^\top \in \mathbb{R}^4 \). We apply epipolar geometry to the projected correspondences, which yields a homogeneous linear system \( \mathbf{A}_i \tilde{\mathbf{X}}_i = \mathbf{0} \), where \( \mathbf{A}_i \in \mathbb{R}^{4 \times 4} \) stacks the constraints from both views. To solve it stably, we rewrite it in the least-squares form:
\begin{equation}
\hat{\mathbf{X}}_i = \arg\min_{\mathbf{X}} \left\| \mathbf{A}_i' \mathbf{X} + \mathbf{b}_i \right\|_2^2,
\end{equation}
where \( \mathbf{A}_i' \in \mathbb{R}^{4 \times 3} \) and \( \mathbf{b}_i \in \mathbb{R}^{4} \) are obtained by separating the last column of \( \mathbf{A}_i \). The triangulated point \( \hat{\mathbf{X}}_i \) is transformed to the central camera frame via extrinsics \( \mathbf{R}_c \in \mathbb{R}^{3 \times 3} \), \( \mathbf{t}_c \in \mathbb{R}^3 \), and the depth prior is defined as the $z$-component:
\begin{equation}
D_{\text{epi}} = \left[ \mathbf{R}_c \hat{\mathbf{X}}_i + \mathbf{t}_c \right]_z.
\end{equation}
After that, we apply an edge-aware Log-L1 loss~\cite{turkulainen2025dn} using \( D_{\text{epi}} \) as follows:
\begin{equation}
L_{\text{epi}}
=
\frac{1}{HW}
\sum_{x,y}
\sum_{k \in \{h, v\}}
\log\!\left( 1 + \big| D_c^{'} - D_{\text{epi}} \big| \right) \cdot
e^{- \bigl| \nabla_{k} I_c \bigr| },
\end{equation}
where \( D_c^{'}\) denotes the depth sampled at the same spatial locations as \( D_{\text{epi}} \). Through this approach, we enforce depth regularization in a self-supervised manner.

\begin{table*}[t]
\centering
\fontsize{50pt}{50pt}\selectfont
\renewcommand{\arraystretch}{1.2}
\resizebox{\textwidth}{!}{
\begin{tabular}{l|ccc|ccc|ccc|ccc|ccc|ccc}
\toprule
\multirow{4}{*}{Method}
& \multicolumn{12}{c|}{SeaThru-NeRF} 
& \multicolumn{6}{c}{In-the-Wild} \\
\cmidrule(lr){2-13} \cmidrule(lr){14-19}
& \multicolumn{3}{c}{\textit{Cura\c{c}ao}} & \multicolumn{3}{c}{\textit{Panama}} 
& \multicolumn{3}{c}{\textit{JapaneseGardens-Redsea}} & \multicolumn{3}{c|}{\textit{IUI3-Redsea}} & \multicolumn{3}{c}{\textit{Coral}} & \multicolumn{3}{c}{\textit{Composite}} \\
\cmidrule(lr){2-4} \cmidrule(lr){5-7} \cmidrule(lr){8-10} \cmidrule(lr){11-13} \cmidrule(lr){14-16} \cmidrule(lr){17-19}
& PSNR~$\uparrow$ & SSIM~$\uparrow$ & LPIPS~$\downarrow$
& PSNR~$\uparrow$ & SSIM~$\uparrow$ & LPIPS~$\downarrow$
& PSNR~$\uparrow$ & SSIM~$\uparrow$ & LPIPS~$\downarrow$
& PSNR~$\uparrow$ & SSIM~$\uparrow$ & LPIPS~$\downarrow$
& PSNR~$\uparrow$ & SSIM~$\uparrow$ & LPIPS~$\downarrow$
& PSNR~$\uparrow$ & SSIM~$\uparrow$ & LPIPS~$\downarrow$\\
\midrule
$\text{SeaThru-NeRF}$
& \cellcolor{yellow!20}{30.86} & 0.869 & 0.216
& 28.41 & 0.833 & 0.219
& 22.34 & 0.763 & 0.263
& 27.27 & 0.791 & 0.293
& 25.28 & 0.732 & 0.315
& 24.15 & 0.722 & 0.315 \\
$\text{SeaThru-NeRF-NS}$
& 28.58 & \cellcolor{yellow!20}{0.907} & \cellcolor{yellow!20}{0.145}
& \cellcolor{yellow!20}{30.60} & \cellcolor{yellow!20}{0.935} & \cellcolor{yellow!20}{0.086}
& \cellcolor{yellow!20}{23.11} & 0.866 & \cellcolor{yellow!20}{0.148}
& \cellcolor{yellow!20}{29.00} & \cellcolor{yellow!20}{0.885} & \cellcolor{red!20}{0.133}
& 27.60 & 0.882 & \cellcolor{yellow!20}{0.119}
& 20.14 & 0.732 & 0.338 \\
$\text{3DGS}$
& 28.92 & 0.881 & 0.202
& 29.35 & 0.899 & 0.137
& 21.03 & 0.859 & 0.205
& 22.48 & 0.832 & 0.258
& \cellcolor{orange!20}{29.11} & \cellcolor{orange!20}{0.906} & \cellcolor{orange!20}{0.112}
& \cellcolor{red!20}{26.94} & \cellcolor{orange!20}{0.877} & \cellcolor{yellow!20}{0.164} \\
$\text{SeaSplat}$
& 29.77 & 0.898 & 0.172
& 28.65 & 0.904 & 0.114
& 23.07 & \cellcolor{yellow!20}{0.875} & 0.155
& 27.23 & 0.868 & \cellcolor{orange!20}{0.183}
& \cellcolor{yellow!20}{28.41} & \cellcolor{yellow!20}{0.897} & 0.125
& \cellcolor{yellow!20}{26.22} & \cellcolor{yellow!20}{0.866} & \cellcolor{orange!20}{0.156} \\
$\text{WaterSplatting}$
& \cellcolor{orange!20}{32.32} & \cellcolor{orange!20}{0.954} & \cellcolor{orange!20}{0.119}
& \cellcolor{orange!20}{31.71} & \cellcolor{orange!20}{0.946} & \cellcolor{orange!20}{0.078}
& \cellcolor{orange!20}{24.77} & \cellcolor{orange!20}{0.897} & \cellcolor{orange!20}{0.119}
& \cellcolor{orange!20}{29.84} & \cellcolor{orange!20}{0.890} & \cellcolor{yellow!20}{0.201}
& 28.19 & 0.877 & 0.160
& 25.47 & 0.849 & 0.177 \\
\textbf{Ours}
& \cellcolor{red!20}{34.56} & \cellcolor{red!20}{0.961} & \cellcolor{red!20}{0.113}
& \cellcolor{red!20}{32.74} & \cellcolor{red!20}{0.957} & \cellcolor{red!20}{0.072}
& \cellcolor{red!20}{25.35} & \cellcolor{red!20}{0.908} & \cellcolor{red!20}{0.114}
& \cellcolor{red!20}{30.17} & \cellcolor{red!20}{0.913} & \cellcolor{orange!20}{0.183}
& \cellcolor{red!20}{29.15} & \cellcolor{red!20}{0.909} & \cellcolor{red!20}{0.105}
& \cellcolor{orange!20}{26.39} & \cellcolor{red!20}{0.883} & \cellcolor{red!20}{0.128} \\
\bottomrule
\end{tabular}
}
\caption{Quantitative evaluation of novel view synthesis performance on real-world underwater scenes compared with reproducible prior methods. \colorbox{red!20}{\textit{1st}}, \colorbox{orange!20}{\textit{2nd}}, and \colorbox{yellow!20}{\textit{3rd}} indicate performance ranking.}
\label{table:main1}
\end{table*}

\begin{table}[t]
\centering
\fontsize{20pt}{20pt}\selectfont
\renewcommand{\arraystretch}{1.1}
\resizebox{\columnwidth}{!}{
\begin{tabular}{l|ccc|ccc}
\toprule
\multirow{2.4}{*}{Method}
& \multicolumn{3}{c}{\textit{Underwater}} & \multicolumn{3}{c}{\textit{Foggy}} \\
\cmidrule(lr){2-4} \cmidrule(lr){5-7}
& PSNR~$\uparrow$ & SSIM~$\uparrow$ & LPIPS~$\downarrow$
& PSNR~$\uparrow$ & SSIM~$\uparrow$ & LPIPS~$\downarrow$ \\
\midrule
& \multicolumn{6}{c}{\textbf{Novel View Synthesis}} \\
\cmidrule(lr){2-7}
$\text{SeaThru-NeRF-NS}$
& \cellcolor{yellow!20}{18.94} & 0.669 & 0.359
& 23.04 & 0.724 & 0.280\\
$\text{SeaSplat}$
& 15.62 & \cellcolor{yellow!20}{0.750} & \cellcolor{yellow!20}{0.247}
& \cellcolor{yellow!20}{27.52} & \cellcolor{yellow!20}{0.831} & \cellcolor{yellow!20}{0.137}\\
$\text{WaterSplatting}$
& \cellcolor{orange!20}{28.12} & \cellcolor{orange!20}{0.858} & \cellcolor{orange!20}{0.094}
& \cellcolor{orange!20}{28.45} & \cellcolor{orange!20}{0.860} & \cellcolor{orange!20}{0.097}\\
\textbf{Ours}
& \cellcolor{red!20}{28.80} & \cellcolor{red!20}{0.871} & \cellcolor{red!20}{0.085}
& \cellcolor{red!20}{29.12} & \cellcolor{red!20}{0.873} & \cellcolor{red!20}{0.089}\\
\midrule
& \multicolumn{6}{c}{\textbf{Scene Restoration}} \\
\cmidrule(lr){2-7}
$\text{SeaThru-NeRF-NS}$
& 12.54 & 0.544 & 0.400
& 14.87 & 0.605 & 0.303\\
$\text{SeaSplat}$
& \cellcolor{yellow!20}{17.15} & \cellcolor{yellow!20}{0.719} & \cellcolor{yellow!20}{0.179}
& \cellcolor{orange!20}{19.84} & \cellcolor{yellow!20}{0.744} & \cellcolor{yellow!20}{0.153} \\
$\text{WaterSplatting}$
& \cellcolor{red!20}{18.39} & \cellcolor{orange!20}{0.748} & \cellcolor{orange!20}{0.141} 
& \cellcolor{yellow!20}{19.40} & \cellcolor{orange!20}{0.770} & \cellcolor{orange!20}{0.129} \\
\textbf{Ours}
& \cellcolor{orange!20}{17.16} & \cellcolor{red!20}{0.768} & \cellcolor{red!20}{0.138} 
& \cellcolor{red!20}{20.17} & \cellcolor{red!20}{0.791} & \cellcolor{red!20}{0.114} \\
\bottomrule
\end{tabular}
}
\caption{Quantitative evaluation of novel view synthesis and scene restoration performance on simulated scenes.}
\label{tab:sim_performance}
\end{table}

\subsection{Depth Residual Loss}
Overly dispersed 3D Gaussians along a camera ray may cause those far from the actual surface to appear as floating artifacts in novel views. To alleviate this issue, we apply a depth residual loss that aligns the rendered depth with the $z$-component of individual 3D Gaussians. This residual loss is computed as follows:
\begin{equation}
    L_{\text{res}} = \frac{1}{N'} \sum_{i=1}^{N'} \left| D_c(\mathbf{x}_i) - z_i \right|,
\end{equation}
where \( N' \) denotes the number of 3D Gaussians located within the camera view frustum, \( D_c(\mathbf{x}_i) \) is the rendered depth via alpha-blending at pixel location \( \mathbf{x}_i \), and \( z_i \) is the camera-space $z$-component of the corresponding 3D Gaussian.

\subsection{Depth-aware Alpha Adjustment}
In scattering media, misplaced 3D Gaussians acquire medium-colored contributions that appear as floating artifacts in novel views. To tackle this challenge, we introduce a depth-aware alpha adjustment. During the adjustment stage $t < t_\alpha$, the adjusted opacity $\alpha'_i$ is used for rasterization and is formulated as a weighted sum of the original opacity $\alpha_i$ and the depth-aware opacity $\alpha^d_i$ as follows:
\begin{equation}
\alpha'_i = (1 - w)\,\alpha_i + w\,\alpha^d_i,
\qquad
\alpha^d_i = \phi_{\alpha}(\alpha_i, z_i, \vec{\mathbf{v}}_i),
\end{equation}
where $\phi_{\alpha}$ is the MLP and $\vec{\mathbf{v}}_i$ is the viewing direction vector. Note that, after transition step $t_\alpha$, the weight $w$ is decayed to zero to eliminate inference-time overhead. This approach adaptively adjusts the transparency of each 3D Gaussian using depth and viewing direction cues, suppresses their opacity in scattering-heavy directions, and promotes the pruning of suboptimal 3D Gaussians.

\subsection{Training Objective}
We adopt the photometric loss as follows:
\begin{equation}
    L_{\text{photo}} = (1 - \lambda_{\text{s}})L_{\text{R-L1}}(I_{c}, I_{gt}) + \lambda_{\text{s}} L_{\text{R-SSIM}}(I_{c}, I_{gt}),
    \label{eq:noisegs_loss}
\end{equation}
where \( L_{\text{R-SSIM}} \) and \( \lambda_{\text{s}} \) denote the Regularized SSIM loss~\cite{li2024watersplatting} and its corresponding weight, respectively, with \( \lambda_{\text{s}} = 0.2 \). In addition to the baseline, we incorporate our proposed components into the final training objective:
\begin{equation}
    L_{\text{total}} = L_{\text{photo}} + \lambda_{\text{tri}}L_{\text{tri}} + \lambda_{\text{epi}}L_{\text{epi}} + \lambda_{\text{res}}L_{\text{res}},
\end{equation}
where $\lambda_{\text{tri}}$, $\lambda_{\text{epi}}$, and $\lambda_{\text{res}}$ are the weights of the proposed loss terms. We empirically set $\lambda_{\text{tri}} = 0.1$ and $\lambda_{\text{res}} = 0.01$, while $\lambda_{\text{epi}}$ is annealed from $0.4$ to $0.2$.

\section{Experiment}
\label{sec:exp}
\subsection{Datasets and Metrics}
We evaluate underwater scene reconstruction on four scenes from SeaThru-NeRF dataset~\cite{levy2023seathru} and two scenes from In-the-Wild dataset~\cite{tang2024neural}. In addition, to evaluate scene restoration performance, we generate simulated underwater and foggy scenes using the \textit{Fern} scene from the LLFF dataset~\cite{mildenhall2019local}. We simulate underwater scenes following previous methods~\cite{levy2023seathru} using normalized depth obtained from a depth foundation model~\cite{yang2024depth} with $\beta_D = [1.3,\ 1.2,\ 0.9]$, $\beta_B = [0.95,\ 0.85,\ 0.7]$, and $\beta_\infty = [0.07,\ 0.2,\ 0.39]$. For foggy scenes, we use $\beta_D = [0.5,\ 0.5,\ 0.5]$ and $\beta_B=\beta_\infty=1.2$. To validate our method, we compare scene quality using PSNR, SSIM~\cite{wang2004image}, and LPIPS~\cite{zhang2018unreasonable}.

\subsection{Implementation Details}
Our method builds on Nerfstudio~\cite{tancik2023nerfstudio}, and employs a TCNN-based MLP~\cite{tiny-cuda-nn}. We use 7K/3K training steps for densification/finetuning on SeaThru-NeRF~\cite{levy2023seathru} and simulated datasets, and 10K/5K steps on In-the-Wild~\cite{tang2024neural}. At each iteration, $b_v$ is sampled from $[-0.4, 0.4]$, and $b_h$ is set to $1.5b_v$. We enable $L_{\text{tri}}$ and $L_{\text{epi}}$ during iterations 4K–8K and 6K–12K for 10K and 15K training steps, respectively.
The $\tau_{\alpha}$ for synthetic epipolar depth is set to 0.8.
The input resolution is increased from \( \times 1/4 \) to \( \times 1/2 \) at \( 1/5 \) of total training steps, and from \( \times 1/2 \) to full resolution at \( 2/5 \) to boost the training process.
For depth-aware alpha adjustment, \(w\) is set to 0.5 for \(t<t_\alpha\) and is decayed to 0 at \( t_\alpha \). We set the transition step $t_{\alpha}$ to 4K and 10K for 10K and 15K training steps, respectively. All experiments use a single RTX A6000.

\begin{figure}[t!]
  \centering
  \renewcommand{\arraystretch}{0.5}
  \setlength{\tabcolsep}{2pt}
  \newcolumntype{M}[1]{>{\centering\arraybackslash}m{#1}}
  \resizebox{\linewidth}{!}{%
    {\large
    \begin{tabular}{M{0.3cm} M{0.4\linewidth} M{0.4\linewidth} M{0.4\linewidth} M{0.4\linewidth}}
      & \text{UW Rendered} & \text{UW Restored} & \text{Fog Rendered} & \text{Fog Restored} \\
      \rotatebox{90}{\textbf{Ours}} &
      \includegraphics[width=\linewidth]{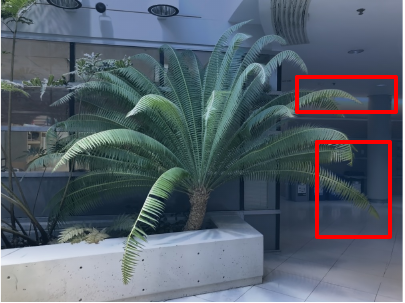} &
      \includegraphics[width=\linewidth]{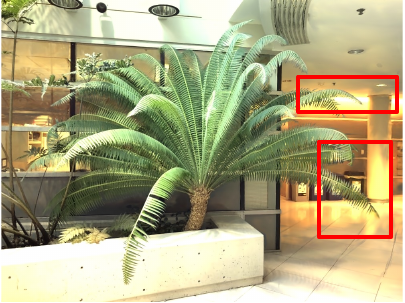} &
      \includegraphics[width=\linewidth]{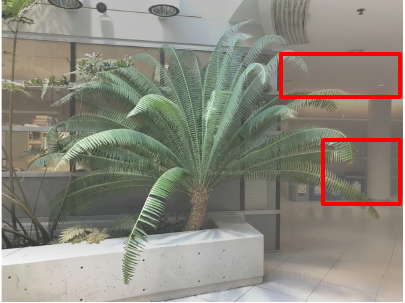} &
      \includegraphics[width=\linewidth]{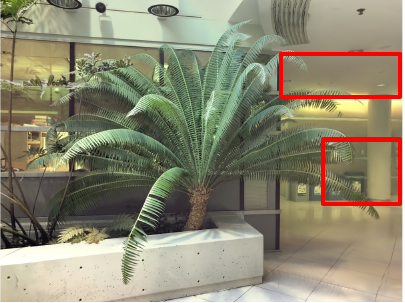} \\
      \rotatebox{90}{\text{WaterSplatting}} &
      \includegraphics[width=\linewidth]{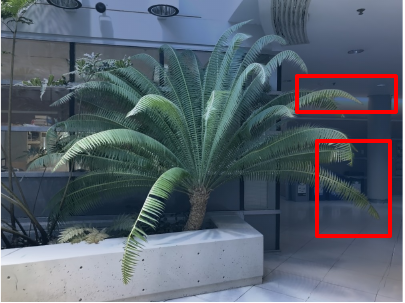} &
      \includegraphics[width=\linewidth]{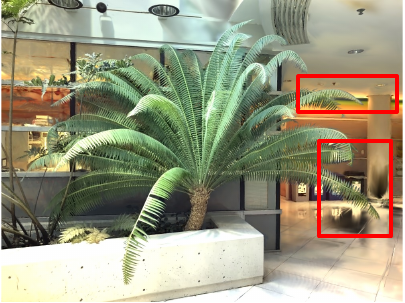} &
      \includegraphics[width=\linewidth]{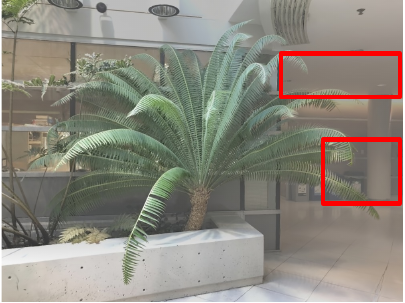} &
      \includegraphics[width=\linewidth]{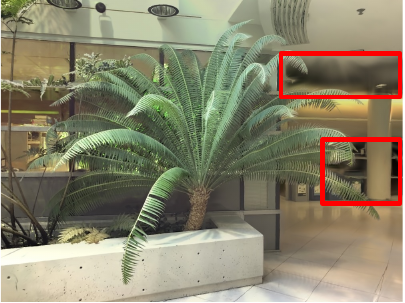} \\
      \rotatebox{90}{\text{SeaSplat}} &
      \includegraphics[width=\linewidth]{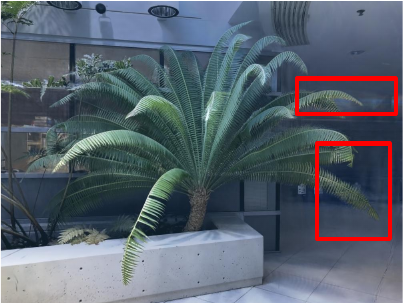} &
      \includegraphics[width=\linewidth]{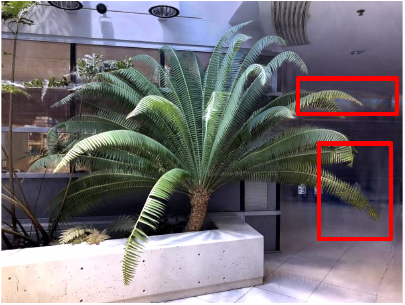} &
      \includegraphics[width=\linewidth]{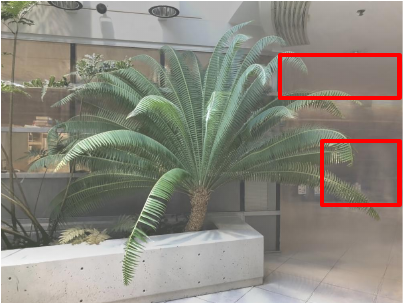} &
      \includegraphics[width=\linewidth]{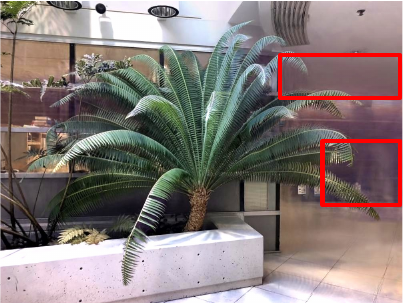} \\
    \end{tabular}
    }
  }
  \caption{Qualitative comparison of scene reconstruction and restoration in underwater and foggy environments.}
  \label{fig:sim_uw_restoration}
\end{figure}

\subsection{Performance Comparison}
\subsubsection{Real-world Underwater Scenes}
We evaluate OceanSplat with previous methods, including the official and Nerfstudio versions of SeaThru-NeRF~\cite{levy2023seathru}, as well as 3DGS~\cite{kerbl20233d}, SeaSplat~\cite{yang2024seasplat}, and WaterSplatting~\cite{li2024watersplatting}. All results were reproduced using the publicly released code.
As shown in \Tabref{table:main1}, OceanSplat achieves superior performance in novel view synthesis across most real-world underwater scenes. On the SeaThru-NeRF dataset~\cite{levy2023seathru}, OceanSplat outperforms the previous state-of-the-art, WaterSplatting~\cite{li2024watersplatting}, and SeaThru-NeRF-NS~\cite{levy2023seathru} by average PSNR margins of 1.05 dB and 2.88 dB, respectively. On the In-the-Wild dataset~\cite{tang2024neural}, OceanSplat reduces the average LPIPS by $0.021$ compared to 3DGS and surpasses SeaSplat~\cite{yang2024seasplat} with average gains of $0.46$ dB in PSNR and $0.014$ in SSIM, along with a $0.024$ reduction in LPIPS.

\Figref{fig:qualitative_comparison} compares RGB and depth renderings across three underwater scenes. In the \textit{Cura\c{c}ao} scene, prior methods suffer from degraded reconstructions due to floating artifacts, whereas our method models the 3D scene with significantly fewer floating artifacts and accurate object representation. In \textit{J.G-Redsea} scene, prior methods yield plausible RGB images but their depth maps reveal erroneous 3D Gaussians in water regions. In contrast, our method maintains consistent geometry throughout. In the \textit{Coral} scene, most prior methods~\cite{levy2023seathru,kerbl20233d,yang2024seasplat} fail to capture fine-grained coral structures, and WaterSplatting~\cite{li2024watersplatting} shows geometry loss and depth holes, while ours reconstructs a faithful 3D scene.

We additionally compare training time, inference FPS, and VRAM usage with prior methods. As shown in \Tabref{tab:cost}, our method trains faster and consumes less memory than SeaThru-NeRF~\cite{levy2023seathru}, while remaining competitive among 3DGS-based approaches. Compared to WaterSplatting~\cite{li2024watersplatting}, our method requires more rasterizations per iteration and least-squares solving for the synthetic epipolar depth prior, resulting in longer training time per-iteration. Nevertheless, it trains over $\times2$ faster than SeaSplat~\cite{yang2024seasplat} while retaining comparable inference speed to other 3DGS-based methods.

\begin{table}[t]
\centering
\fontsize{10pt}{10pt}\selectfont
\renewcommand{\arraystretch}{1.2}
\setlength{\tabcolsep}{10pt}
\resizebox{\linewidth}{!}{
\begin{tabular}{l|ccc}
\toprule
Method & Training~$\downarrow$ & FPS~$\uparrow$ & Memory (GB)~$\downarrow$ \\
\midrule
SeaThru-NeRF & 18 h 25 m & 0.10 & 37.7  \\
SeaThru-NeRF-NS & 3 h 21 m & 0.29 & 15.1 \\
3DGS & 22 m & 150.28 & 3.7 \\
SeaSplat & 51 m  & 90.05 & 6.0  \\
WaterSplatting & 10 m  & 88.19 & 5.8  \\
\textbf{Ours} & 19 m & 85.67 & 7.6  \\
\bottomrule
\end{tabular}
}
\caption{Comparison of training time, rendering speed (FPS), and memory consumption with prior methods.}
\label{tab:cost}
\end{table}

\subsubsection{Simulated Scattering Scenes}
We compare the scene reconstruction and restoration performance of SeaThru-NeRF-NS~\cite{levy2023seathru}, SeaSplat~\cite{yang2024seasplat}, WaterSplatting~\cite{li2024watersplatting}, and our proposed OceanSplat using a simulated dataset. As shown in \Tabref{tab:sim_performance}, for the novel view synthesis task, OceanSplat achieves an average PSNR improvement of 7.39 dB over SeaSplat~\cite{yang2024seasplat} and 0.67 dB over WaterSplatting~\cite{li2024watersplatting}. For scene restoration, our method also demonstrates superior performance, yielding average SSIM gains of 0.048 over SeaSplat~\cite{yang2024seasplat} and 0.021 over WaterSplatting~\cite{li2024watersplatting}.
\Figref{fig:sim_uw_restoration} shows that, in both simulated scenes, previous methods assign 3D Gaussians to the scattering medium (e.g., ceilings or walls), resulting in restoration holes, whereas our method accurately represents scene structures in both underwater and foggy scenes.

\begin{table}[t]
\centering
\fontsize{10pt}{10pt}\selectfont
\renewcommand{\arraystretch}{1.2}
\setlength{\tabcolsep}{10pt}
\resizebox{0.8\linewidth}{!}{
\begin{tabular}{l|ccc}
\toprule
Configuration & PSNR~$\uparrow$ & SSIM~$\uparrow$ & LPIPS~$\downarrow$ \\
\midrule
\textbf{Full Model} & \cellcolor{red!20}{34.56} & \cellcolor{red!20}{0.961} & \cellcolor{red!20}{0.113} \\
w/o $L_{res}$ & \cellcolor{orange!20}{34.30} & \cellcolor{orange!20}{0.960} & \cellcolor{orange!20}{0.115}  \\
w/o $L_{epi}$ & 33.82 & \cellcolor{yellow!20}{0.959} & 0.120 \\
w/o $L_{tri}$ & 33.20 & 0.957 & \cellcolor{orange!20}{0.115} \\
w/o $\alpha^{d}$ & \cellcolor{yellow!20}{33.90} & \cellcolor{orange!20}{0.960} & \cellcolor{yellow!20}{0.116} \\
\bottomrule
\end{tabular}
}
\caption{Ablation study demonstrating the effectiveness of each component in our method.}
\label{tab:ablation_effect}
\end{table}

\begin{table}[t]
\centering
\fontsize{10pt}{10pt}\selectfont
\renewcommand{\arraystretch}{1.2}
\resizebox{\linewidth}{!}{
\begin{tabular}{l@{\hspace{-4pt}}l|ccc|c}
\toprule
Configuration
& & PSNR~$\uparrow$ & SSIM~$\uparrow$ & LPIPS~$\downarrow$ & Training~$\downarrow$ \\
\midrule
Baseline & $(P_c)$ & 32.20 & \cellcolor{yellow!20}{0.954} & 0.120 & 9m  \\
Binocular & $(P_c+P_h)$ & \cellcolor{yellow!20}{33.62} & \cellcolor{orange!20}{0.957} & \cellcolor{yellow!20}{0.119} & 13m \\
Trinocular & $(P_c+P_h+P_{h'})$ & \cellcolor{orange!20}{33.72} & \cellcolor{orange!20}{0.957} & \cellcolor{orange!20}{0.118} & 16m \\
\textbf{Trinocular} & $(P_c+P_h+P_v)$ & \cellcolor{red!20}{33.91} & \cellcolor{red!20}{0.958} & \cellcolor{red!20}{0.115} & 16m \\
\bottomrule
\end{tabular}
}
\caption{Ablation study on stereo configurations, comparing the impact of horizontally and vertically translated camera poses on performance and training time.}
\label{tab:ablation_design}
\end{table}

\subsection{Ablation Study}
In \Tabref{tab:ablation_effect}, we evaluate the effectiveness of each component on the \textit{Cura\c{c}ao} scene. Removing $L_{\text{tri}}$ yields the largest PSNR drop, confirming that trinocular view consistency guides spatial optimization of 3D Gaussians, enabling structurally coherent novel view synthesis. Excluding $L_{\text{epi}}$ significantly increases LPIPS, validating our self-supervised depth regularization in underwater scenes with a lack of geometric cues. Omitting $L_{\text{res}}$ and depth-aware opacity $\alpha^{d}$ degrades overall quality, highlighting their role in mitigating medium entanglement through reduced depth ambiguity.

\Tabref{tab:ablation_design} compares rendering performance and training time across different stereo configurations, including the baseline~\cite{li2024watersplatting}. Binocular stereo setup $(P_c+P_h)$ improves over baseline, while adding another horizontal stereo view $(P_c+P_h+P_{h'})$ provides minimal gain despite longer training. In contrast, introducing vertical disparity $(P_c+P_h+P_v)$ achieves a $0.29$ dB increase in PSNR over the binocular setup. This shows that using horizontally and vertically translated camera poses enhances  performance without compromising training time, as shown in \Figref{fig:trinocular_1}.

\section{Conclusion}
\label{sec:conclusion}
We introduce OceanSplat, a novel framework for enhancing geometric consistency in underwater scene reconstruction. Primarily, we impose trinocular view consistency across horizontally and vertically translated virtual viewpoints, effectively aligning 3D Gaussians with scene structure. In addition, we introduce synthetic epipolar depth priors derived via triangulation of correspondences from the translated viewpoints, enabling self-supervised depth regularization. Lastly, we propose a depth-aware alpha adjustment that uses depth and viewing direction to suppress medium-induced floating artifacts. Through these contributions, OceanSplat achieves high-fidelity underwater scene reconstruction. Future work will focus on addressing non-rigid object representations and eliminating the dependence on SfM, thereby broadening the applicability to dynamic underwater scenes.

\section{Acknowledgments}
This work was supported in part by the National Research Foundation of Korea (NRF) grant funded by the Korea government (MSIT) (RS-2023-00217689, 50\%; RS-2024-00358935, 40\%), and in part by Institute of Information \& communications Technology Planning \& Evaluation (IITP) under the Artificial Intelligence Convergence Innovation Human Resources Development grant funded by the Korea government (MSIT) (IITP-2026-RS-2023-00254177, 10\%). Minseong Kweon acknowledges travel support from the Minnesota Robotics Institute (MnRI).

\begin{small}
\bibliography{aaai2026}

@inproceedings{mildenhall2020nerf,
 title={NeRF: Representing Scenes as Neural Radiance Fields for View Synthesis},
 author={Ben Mildenhall and Pratul P. Srinivasan and Matthew Tancik and Jonathan T. Barron and Ravi Ramamoorthi and Ren Ng},
 year={2020},
 booktitle={Proceedings of the European Conference on Computer Vision},
}

@article{kerbl20233d,
  title={3d gaussian splatting for real-time radiance field rendering.},
  author={Kerbl, Bernhard and Kopanas, Georgios and Leimk{\"u}hler, Thomas and Drettakis, George},
  journal={ACM Transactions on Graphics},
  volume={42},
  number={4},
  pages={139--1},
  year={2023}
}

@inproceedings{sethuraman2023waternerf,
  title={Waternerf: Neural radiance fields for underwater scenes},
  author={Sethuraman, Advaith Venkatramanan and Ramanagopal, Manikandasriram Srinivasan and Skinner, Katherine A},
  booktitle={OCEANS 2023-MTS/IEEE US Gulf Coast},
  pages={1--7},
  year={2023},
  organization={IEEE}
}

@article{zhang2023beyond,
  title={Beyond nerf underwater: Learning neural reflectance fields for true color correction of marine imagery},
  author={Zhang, Tianyi and Johnson-Roberson, Matthew},
  journal={IEEE Robotics and Automation Letters},
  volume={8},
  number={10},
  pages={6467--6474},
  year={2023},
  publisher={IEEE}
}

@inproceedings{levy2023seathru,
  title={Seathru-nerf: Neural radiance fields in scattering media},
  author={Levy, Deborah and Peleg, Amit and Pearl, Naama and Rosenbaum, Dan and Akkaynak, Derya and Korman, Simon and Treibitz, Tali},
  booktitle={Proceedings of the IEEE/CVF Conference on Computer Vision and Pattern Recognition},
  pages={56--65},
  year={2023}
}

@inproceedings{tang2024neural,
  title={Neural underwater scene representation},
  author={Tang, Yunkai and Zhu, Chengxuan and Wan, Renjie and Xu, Chao and Shi, Boxin},
  booktitle={Proceedings of the IEEE/CVF Conference on Computer Vision and Pattern Recognition},
  pages={11780--11789},
  year={2024}
}

@article{wang2024uw,
  title={UW-GS: Distractor-aware 3d gaussian splatting for enhanced underwater scene reconstruction},
  author={Wang, Haoran and Anantrasirichai, Nantheera and Zhang, Fan and Bull, David},
  journal={arXiv preprint arXiv:2410.01517},
  year={2024}
}

@article{li2024watersplatting,
  title={Watersplatting: Fast underwater 3d scene reconstruction using gaussian splatting},
  author={Li, Huapeng and Song, Wenxuan and Xu, Tianao and Elsig, Alexandre and Kulhanek, Jonas},
  journal={arXiv preprint arXiv:2408.08206},
  year={2024}
}

@article{yang2024seasplat,
  title={Seasplat: Representing underwater scenes with 3d gaussian splatting and a physically grounded image formation model},
  author={Yang, Daniel and Leonard, John J and Girdhar, Yogesh},
  journal={arXiv preprint arXiv:2409.17345},
  year={2024}
}

@inproceedings{akkaynak2018revised,
  title={A revised underwater image formation model},
  author={Akkaynak, Derya and Treibitz, Tali},
  booktitle={Proceedings of the IEEE/CVF Conference on Computer Vision and Pattern Recognition},
  pages={6723--6732},
  year={2018}
}

@inproceedings{akkaynak2017space,
  title={What is the space of attenuation coefficients in underwater computer vision?},
  author={Akkaynak, Derya and Treibitz, Tali and Shlesinger, Tom and Loya, Yossi and Tamir, Raz and Iluz, David},
  booktitle={Proceedings of the IEEE/CVF Conference on Computer Vision and Pattern Recognition},
  pages={4931--4940},
  year={2017}
}

@inproceedings{akkaynak2019sea,
  title={Sea-thru: A method for removing water from underwater images},
  author={Akkaynak, Derya and Treibitz, Tali},
  booktitle={Proceedings of the IEEE/CVF Conference on Computer Vision and Pattern Recognition},
  pages={1682--1691},
  year={2019}
}

@inproceedings{ramazzina2023scatternerf,
  title={Scatternerf: Seeing through fog with physically-based inverse neural rendering},
  author={Ramazzina, Andrea and Bijelic, Mario and Walz, Stefanie and Sanvito, Alessandro and Scheuble, Dominik and Heide, Felix},
  booktitle={Proceedings of the IEEE/CVF International Conference on Computer Vision},
  pages={17957--17968},
  year={2023}
}

@inproceedings{li2024dngaussian,
  title={Dngaussian: Optimizing sparse-view 3d gaussian radiance fields with global-local depth normalization},
  author={Li, Jiahe and Zhang, Jiawei and Bai, Xiao and Zheng, Jin and Ning, Xin and Zhou, Jun and Gu, Lin},
  booktitle={Proceedings of the IEEE/CVF Conference on Computer Vision and Pattern Recognition},
  pages={20775--20785},
  year={2024}
}

@article{safadoust2024self,
  title={Self-Evolving Depth-Supervised 3D Gaussian Splatting from Rendered Stereo Pairs},
  author={Safadoust, Sadra and Tosi, Fabio and G{\"u}ney, Fatma and Poggi, Matteo},
  journal={arXiv preprint arXiv:2409.07456},
  year={2024}
}

@article{han2024binocular,
  title={Binocular-guided 3d gaussian splatting with view consistency for sparse view synthesis},
  author={Han, Liang and Zhou, Junsheng and Liu, Yu-Shen and Han, Zhizhong},
  journal={arXiv preprint arXiv:2410.18822},
  year={2024}
}

@inproceedings{chung2024depth,
  title={Depth-regularized optimization for 3d gaussian splatting in few-shot images},
  author={Chung, Jaeyoung and Oh, Jeongtaek and Lee, Kyoung Mu},
  booktitle={Proceedings of the IEEE/CVF Conference on Computer Vision and Pattern Recognition},
  pages={811--820},
  year={2024}
}

@inproceedings{schonberger2016structure,
  title={Structure-from-motion revisited},
  author={Schonberger, Johannes L and Frahm, Jan-Michael},
  booktitle={Proceedings of the IEEE/CVF Conference on Computer Vision and Pattern Recognition},
  pages={4104--4113},
  year={2016}
}

@article{wang2004image,
  title={Image quality assessment: from error visibility to structural similarity},
  author={Wang, Zhou and Bovik, Alan C and Sheikh, Hamid R and Simoncelli, Eero P},
  journal={IEEE Transactions on Image Processing},
  volume={13},
  number={4},
  pages={600--612},
  year={2004},
  publisher={IEEE}
}

@inproceedings{beall20103d,
  title={3D reconstruction of underwater structures},
  author={Beall, Chris and Lawrence, Brian J and Ila, Viorela and Dellaert, Frank},
  booktitle={2010 IEEE/RSJ International Conference on Intelligent Robots and Systems},
  pages={4418--4423},
  year={2010},
  organization={IEEE}
}

@inproceedings{zhang2018unreasonable,
  title={The unreasonable effectiveness of deep features as a perceptual metric},
  author={Zhang, Richard and Isola, Phillip and Efros, Alexei A and Shechtman, Eli and Wang, Oliver},
  booktitle={Proceedings of the IEEE/CVF Conference on Computer Vision and Pattern Recognition},
  pages={586--595},
  year={2018}
}

@misc{tiny-cuda-nn,
	author = {M\"uller, Thomas},
	license = {BSD-3-Clause},
	month = {4},
	title = {{tiny-cuda-nn}},
	url = {https://github.com/NVlabs/tiny-cuda-nn},
	version = {2.0},
	year = {2021}
}

@article{mildenhall2019local,
  title={Local light field fusion: Practical view synthesis with prescriptive sampling guidelines},
  author={Mildenhall, Ben and Srinivasan, Pratul P and Ortiz-Cayon, Rodrigo and Kalantari, Nima Khademi and Ramamoorthi, Ravi and Ng, Ren and Kar, Abhishek},
  journal={ACM Transactions on Graphics},
  volume={38},
  number={4},
  pages={1--14},
  year={2019},
  publisher={ACM New York, NY, USA}
}

@inproceedings{yang2024depth,
  title={Depth anything: Unleashing the power of large-scale unlabeled data},
  author={Yang, Lihe and Kang, Bingyi and Huang, Zilong and Xu, Xiaogang and Feng, Jiashi and Zhao, Hengshuang},
  booktitle={Proceedings of the IEEE/CVF Conference on Computer Vision and Pattern Recognition},
  pages={10371--10381},
  year={2024}
}

@inproceedings{tancik2023nerfstudio,
  title={Nerfstudio: A modular framework for neural radiance field development},
  author={Tancik, Matthew and Weber, Ethan and Ng, Evonne and Li, Ruilong and Yi, Brent and Wang, Terrance and Kristoffersen, Alexander and Austin, Jake and Salahi, Kamyar and Ahuja, Abhik and others},
  booktitle={ACM SIGGRAPH 2023 Conference Papers},
  pages={1--12},
  year={2023}
}

@inproceedings{mildenhall2022nerf,
  title={Nerf in the dark: High dynamic range view synthesis from noisy raw images},
  author={Mildenhall, Ben and Hedman, Peter and Martin-Brualla, Ricardo and Srinivasan, Pratul P and Barron, Jonathan T},
  booktitle={Proceedings of the IEEE/CVF Conference on Computer Vision and Pattern Recognition},
  pages={16190--16199},
  year={2022}
}

@inproceedings{marques2020l2uwe,
  title={L2uwe: A framework for the efficient enhancement of low-light underwater images using local contrast and multi-scale fusion},
  author={Marques, Tunai Porto and Albu, Alexandra Branzan},
  booktitle={Proceedings of the IEEE/CVF Conference on Computer Vision and Pattern Recognition workshops},
  pages={538--539},
  year={2020}
}

@article{wynn2014autonomous,
  title={Autonomous Underwater Vehicles (AUVs): Their past, present and future contributions to the advancement of marine geoscience},
  author={Wynn, Russell B and Huvenne, Veerle AI and Le Bas, Timothy P and Murton, Bramley J and Connelly, Douglas P and Bett, Brian J and Ruhl, Henry A and Morris, Kirsty J and Peakall, Jeffrey and Parsons, Daniel R and others},
  journal={Marine Geology},
  volume={352},
  pages={451--468},
  year={2014},
  publisher={Elsevier}
}

@article{huang2025visual,
  title={Visual enhancement and 3D representation for underwater scenes: a review},
  author={Huang, Guoxi and Wang, Haoran and Seymour, Brett and Kovacs, Evan and Ellerbrock, John and Blackham, Dave and Anantrasirichai, Nantheera},
  journal={arXiv preprint arXiv:2505.01869},
  year={2025}
}

@InProceedings{turkulainen2025dn,
        title={DN-Splatter: Depth and Normal Priors for Gaussian Splatting and Meshing}, 
        author={Matias Turkulainen and Xuqian Ren and Iaroslav Melekhov and Otto Seiskari and Esa Rahtu and Juho Kannala},
        booktitle = {Proceedings of the IEEE/CVF Winter Conference on Applications of Computer Vision},
        year={2025}
}

@inproceedings{de2021impact,
  title={Impact of colour on robustness of deep neural networks},
  author={De, Kanjar and Pedersen, Marius},
  booktitle={Proceedings of the IEEE/CVF International Conference on Computer Vision},
  pages={21--30},
  year={2021}
}

@article{jaffe2002computer,
  title={Computer modeling and the design of optimal underwater imaging systems},
  author={Jaffe, Jules S},
  journal={IEEE Journal of Oceanic Engineering},
  volume={15},
  number={2},
  pages={101--111},
  year={2002},
  publisher={IEEE}
}

@inproceedings{mcglamery1980computer,
  title={A computer model for underwater camera systems},
  author={McGlamery, BL},
  booktitle={Ocean Optics VI},
  volume={208},
  pages={221--231},
  year={1980},
  organization={SPIE}
}

@article{zhang2024dcgf,
  title={DCGF: Diffusion-Color Guided Framework for Underwater Image Enhancement},
  author={Zhang, Yuhan and Yuan, Jieyu and Cai, Zhanchuan},
  journal={IEEE Transactions on Geoscience and Remote Sensing},
  year={2024},
  publisher={IEEE}
}

@article{yu2023task,
  title={Task-friendly underwater image enhancement for machine vision applications},
  author={Yu, Meng and Shen, Liquan and Wang, Zhengyong and Hua, Xia},
  journal={IEEE Transactions on Geoscience and Remote Sensing},
  volume={62},
  pages={1--14},
  year={2023},
  publisher={IEEE}
}

@article{marre2019monitoring,
  title={Monitoring marine habitats with photogrammetry: a cost-effective, accurate, precise and high-resolution reconstruction method},
  author={Marre, Guilhem and Holon, Florian and Luque, Sandra and Boissery, Pierre and Deter, Julie},
  journal={Frontiers in Marine Science},
  volume={6},
  pages={276},
  year={2019},
  publisher={Frontiers Media SA}
}

@inproceedings{girdhar2023curee,
  title={CUREE: A Curious Underwater Robot for Ecosystem Exploration},
  author={Girdhar, Yogesh and McGuire, Nathan and Cai, Levi and Jamieson, Stewart and McCammon, Seth and Claus, Brian and Soucie, John E San and Todd, Jessica E and Mooney, T Aran},
  booktitle={IEEE International Conference on Robotics and Automation},
  pages={11411-11417},
  year={2023},
  organization={IEEE}
}

@inproceedings{joshi2022underwater,
  title={Underwater exploration and mapping},
  author={Joshi, Bharat and Xanthidis, Marios and Roznere, Monika and Burgdorfer, Nathaniel J and Mordohai, Philippos and Li, Alberto Quattrini and Rekleitis, Ioannis},
  booktitle={2022 IEEE/OES Autonomous Underwater Vehicles Symposium},
  pages={1--7},
  year={2022},
  organization={IEEE}
}

@article{kapoutsis2016real,
  title={Real-time adaptive multi-robot exploration with application to underwater map construction},
  author={Kapoutsis, Athanasios Ch and Chatzichristofis, Savvas A and Doitsidis, Lefteris and De Sousa, Joao Borges and Pinto, Jose and Braga, Jose and Kosmatopoulos, Elias B},
  journal={Autonomous Robots},
  volume={40},
  number={6},
  pages={987--1015},
  year={2016},
  publisher={Springer}
}

@article{liu2020underwater,
  title={Underwater hyperspectral imaging technology and its applications for detecting and mapping the seafloor: A review},
  author={Liu, Bohan and Liu, Zhaojun and Men, Shaojie and Li, Yongfu and Ding, Zhongjun and He, Jiahao and Zhao, Zhigang},
  journal={Sensors},
  volume={20},
  number={17},
  pages={4962},
  year={2020},
  publisher={MDPI}
}

@article{fang2023integration,
  title={Integration of ROV and vision-based underwater inspection for Limnoperna fortunei in water conveyance structure},
  author={Fang, Xin and Li, Heng and Zhang, Sherong and Zhang, Jikang and Wang, Chao and Wang, Xiaohua and Ma, Ziao and Jia, He},
  journal={Engineering Applications of Artificial Intelligence},
  volume={124},
  pages={106575},
  year={2023},
  publisher={Elsevier}
}

@article{shamsafar2021tristereonet,
  title={TriStereoNet: A Trinocular Framework for Multi-baseline Disparity Estimation},
  author={Shamsafar, Faranak and Zell, Andreas},
  journal={arXiv preprint arXiv:2111.12502},
  year={2021}
}

@article{okutomi1993multiple,
  title={A multiple-baseline stereo},
  author={Okutomi, Masatoshi and Kanade, Takeo},
  journal={IEEE Transactions on Pattern Analysis and Machine Intelligence},
  volume={15},
  number={4},
  pages={353--363},
  year={1993},
  publisher={IEEE}
}

@inproceedings{imran2020unsupervised,
  title={Unsupervised Monocular Depth Estimation with Multi-Baseline Stereo},
  author={Imran, Saad and Khan, Muhammad Umar Karim and Mukaram, Sikander and Kyung, Chong-Min},
  booktitle={Proceedings of the British Machine Vision Conference},
  year={2020}
}

@inproceedings{skinner2017automatic,
  title={Automatic color correction for 3D reconstruction of underwater scenes},
  author={Skinner, Katherine A and Iscar, Eduardo and Johnson-Roberson, Matthew},
  booktitle={IEEE International Conference on Robotics and Automation},
  pages={5140--5147},
  year={2017},
  organization={IEEE}
}

@inproceedings{debortoli2018real,
  title={Real-time underwater 3D reconstruction using global context and active labeling},
  author={DeBortoli, Robert and Nicolai, Austin and Li, Fuxin and Hollinger, Geoffrey A},
  booktitle={IEEE International Conference on Robotics and Automation},
  pages={6204--6211},
  year={2018},
  organization={IEEE}
}

@inproceedings{zhang2025decoupling,
  title={Decoupling Scattering: Pseudo-Label Guided NeRF for Scenes with Scattering Media},
  author={Zhang, Mingyang and Zhang, Junkang and Fang, Faming and Zhang, Guixu},
  booktitle={Proceedings of the AAAI Conference on Artificial Intelligence},
  year={2025}
}

@article{kutulakos2000theory,
  title={A theory of shape by space carving},
  author={Kutulakos, Kiriakos N and Seitz, Steven M},
  journal={International Journal of Computer Vision},
  volume={38},
  number={3},
  pages={199--218},
  year={2000},
  publisher={Springer}
}

@article{chen2021underwater,
  title={Underwater image enhancement based on deep learning and image formation model},
  author={Chen, Xuelei and Zhang, Pin and Quan, Lingwei and Yi, Chao and Lu, Cunyue},
  journal={arXiv preprint arXiv:2101.00991},
  year={2021}
}

@article{furukawa2009accurate,
  title={Accurate, dense, and robust multiview stereopsis},
  author={Furukawa, Yasutaka and Ponce, Jean},
  journal={IEEE Transactions on Pattern Analysis and Machine Intelligence},
  volume={32},
  number={8},
  pages={1362--1376},
  year={2009},
  publisher={IEEE}
}

@inproceedings{galliani2015massively,
  title={Massively parallel multiview stereopsis by surface normal diffusion},
  author={Galliani, Silvano and Lasinger, Katrin and Schindler, Konrad},
  booktitle={Proceedings of the IEEE/CVF International Conference on Computer Vision},
  pages={873--881},
  year={2015}
}

@inproceedings{yao2018mvsnet,
  title={Mvsnet: Depth inference for unstructured multi-view stereo},
  author={Yao, Yao and Luo, Zixin and Li, Shiwei and Fang, Tian and Quan, Long},
  booktitle={Proceedings of the European Conference on Computer Vision},
  pages={767--783},
  year={2018}
}

@article{wang2021neus,
  title={Neus: Learning neural implicit surfaces by volume rendering for multi-view reconstruction},
  author={Wang, Peng and Liu, Lingjie and Liu, Yuan and Theobalt, Christian and Komura, Taku and Wang, Wenping},
  journal={arXiv preprint arXiv:2106.10689},
  year={2021}
}

@article{fu2022geo,
  title={Geo-neus: Geometry-consistent neural implicit surfaces learning for multi-view reconstruction},
  author={Fu, Qiancheng and Xu, Qingshan and Ong, Yew Soon and Tao, Wenbing},
  journal={Advances in Neural Information Processing Systems},
  volume={35},
  pages={3403--3416},
  year={2022}
}

@inproceedings{chen2021mvsnerf,
  title={Mvsnerf: Fast generalizable radiance field reconstruction from multi-view stereo},
  author={Chen, Anpei and Xu, Zexiang and Zhao, Fuqiang and Zhang, Xiaoshuai and Xiang, Fanbo and Yu, Jingyi and Su, Hao},
  booktitle={Proceedings of the IEEE/CVF International Conference on Computer Vision},
  pages={14124--14133},
  year={2021}
}

@inproceedings{guedon2024sugar,
  title={Sugar: Surface-aligned gaussian splatting for efficient 3d mesh reconstruction and high-quality mesh rendering},
  author={Gu{\'e}don, Antoine and Lepetit, Vincent},
  booktitle={Proceedings of the IEEE/CVF Conference on Computer Vision and Pattern Recognition},
  pages={5354--5363},
  year={2024}
}

@article{chen2023neusg,
  title={Neusg: Neural implicit surface reconstruction with 3d gaussian splatting guidance},
  author={Chen, Hanlin and Li, Chen and Lee, Gim Hee},
  journal={arXiv preprint arXiv:2312.00846},
  year={2023}
}

@inproceedings{huang20242d,
  title={2d gaussian splatting for geometrically accurate radiance fields},
  author={Huang, Binbin and Yu, Zehao and Chen, Anpei and Geiger, Andreas and Gao, Shenghua},
  booktitle={ACM SIGGRAPH 2024 Conference Papers},
  pages={1--11},
  year={2024}
}

@article{chen2024pgsr,
  title={Pgsr: Planar-based gaussian splatting for efficient and high-fidelity surface reconstruction},
  author={Chen, Danpeng and Li, Hai and Ye, Weicai and Wang, Yifan and Xie, Weijian and Zhai, Shangjin and Wang, Nan and Liu, Haomin and Bao, Hujun and Zhang, Guofeng},
  journal={IEEE Transactions on Visualization and Computer Graphics},
  year={2024},
  publisher={IEEE}
}

@inproceedings{liu2024mvsgaussian,
  title={Mvsgaussian: Fast generalizable gaussian splatting reconstruction from multi-view stereo},
  author={Liu, Tianqi and Wang, Guangcong and Hu, Shoukang and Shen, Liao and Ye, Xinyi and Zang, Yuhang and Cao, Zhiguo and Li, Wei and Liu, Ziwei},
  booktitle={Proceedings of the European Conference on Computer Vision},
  pages={37--53},
  year={2024},
  organization={Springer}
}

@inproceedings{chen2024mvsplat,
  title={Mvsplat: Efficient 3d gaussian splatting from sparse multi-view images},
  author={Chen, Yuedong and Xu, Haofei and Zheng, Chuanxia and Zhuang, Bohan and Pollefeys, Marc and Geiger, Andreas and Cham, Tat-Jen and Cai, Jianfei},
  booktitle={Proceedings of the European Conference on Computer Vision},
  pages={370--386},
  year={2024},
  organization={Springer}
}

@inproceedings{wu2025sparis,
    title={Sparis: Neural Implicit Surface Reconstruction of Indoor Scenes from Sparse Views},
    author={Yulun Wu and Han Huang and Wenyuan Zhang and Chao Deng and Ge Gao and Ming Gu and Yu-Shen Liu},
    booktitle={Proceedings of the AAAI Conference on Artificial Intelligence},
    year={2025}
}

@inproceedings{huang2025fatesgs,
    title={FatesGS: Fast and Accurate Sparse-View Surface Reconstruction Using Gaussian Splatting with Depth-Feature Consistency},
    author={Han Huang and Yulun Wu and Chao Deng and Ge Gao and Ming Gu and Yu-Shen Liu},
    booktitle={Proceedings of the AAAI Conference on Artificial Intelligence},
    year={2025}
}
\end{small}

\end{document}